\documentclass[Afour,sageh,times]{sagej}

\usepackage{moreverb,url}
\usepackage{enumitem}
\usepackage{wrapfig}
\usepackage{pifont}
\usepackage[colorlinks,bookmarksopen,bookmarksnumbered,citecolor=red,urlcolor=red]{hyperref}
\usepackage[english]{babel}

\usepackage{listings}
\usepackage[dvipsnames]{xcolor}
\usepackage{algorithm}
\usepackage[noend]{algorithmic}
\usepackage[breakable]{tcolorbox}
\usepackage{multirow}

\newcommand\BibTeX{{\rmfamily B\kern-.05em \textsc{i\kern-.025em b}\kern-.08em
T\kern-.1667em\lower.7ex\hbox{E}\kern-.125emX}}

\def\journalname{The International Journal of Robotics Research}
\def\volumeyear{2024}
\def\volumenumber{yy}
\def\issuenumber{zz}

\usepackage{xspace}

\newcommand{\ie}{\textit{\textrm{i.e.}}}
\newcommand{\eg}{\textit{\textrm{e.g.}}}
\newcommand{\etc}{\textit{etc.}}

\newcommand{\cmark}{\ding{51}}%
\newcommand{\xmark}{\ding{55}}%

\def\graphlong{Open Scene Graph\xspace}
\def\graphshort{OSG\xspace}
\def\enginelong{Mapper}
\def\engineshort{Mapper}
\def\navsys{{OSG Navigator}\xspace}
\def\edgehas{``\texttt{has}''}
\def\edgecontains{``\texttt{contains}''}
\def\edgeconnects{``\texttt{connects to}''}
\def\edgenear{``\texttt{is near}''}

\def\ourspr{OSG-Nav}
\def\oursprgt{OSG-Nav-GT}
\def\hm{HM3D}
\def\hmfull{HM3D-Semantics v0.1}
\def\gibson{Gibson}

\def\vo{v^{o}}

\def\vp{v^{p}}
\def\vi{v^{i}}

\def\objnodelabel{$\vo_{label}$}
\def\objnodeimage{$\vo_{img}$}
\def\objnodeid{$\vo_{id}$}
\def\objnodetext{$\vo_{desc}$}

\def\placenode{$\vp{}$}
\def\placenodeclass{$\vp_{cls}$}
\def\placenodelabel{$\vp_{label}$}
\def\placenodeid{$\vp_{id}$}

\def\absinodelabel{$\vi_{label}$}
\def\absinodeid{$\vi_{id}$}

\def\centroidthresh{$\beta_{pix}$}
\def\bboxthresh{$\beta_{IOU}$}

\def\oostate{\mathcal{S}}
\def\ooaction{\mathcal{A}}
\def\ootrans{\mathcal{T}}
\def\ooreward{\mathcal{G}}

\def\ooobs{\Omega}
\def\ooobsmodel{O}

\def\weblink{\urllink[pre = \bgroup\bf, post = \egroup]}

\usepackage{etoolbox}
\makeatletter
\patchcmd{\hyper@makecurrent}{%
    \ifx\Hy@param\Hy@chapterstring
        \let\Hy@param\Hy@chapapp
    \fi
}{%
    \iftoggle{inappendix}{%true-branch
        \@checkappendixparam{chapter}%
        \@checkappendixparam{section}%
        \@checkappendixparam{subsection}%
        \@checkappendixparam{subsubsection}%
        \@checkappendixparam{paragraph}%
        \@checkappendixparam{subparagraph}%
    }{}%
}{}{\errmessage{failed to patch}}

\newcommand*{\@checkappendixparam}[1]{%
    \def\@checkappendixparamtmp{#1}%
    \ifx\Hy@param\@checkappendixparamtmp
        \let\Hy@param\Hy@appendixstring
    \fi
}
\makeatletter

\newtoggle{inappendix}
\togglefalse{inappendix}

\apptocmd{\appendix}{\toggletrue{inappendix}}{}{\errmessage{failed to patch}}
\newcommand\tcb@cnt@specboxautorefname{\graphshort{} Schema}
\newcommand\tcb@cnt@llmpromptboxautorefname{LLM Prompt}
\newcommand\tcb@cnt@vqapromptboxautorefname{VQA Prompt}

\tcbset{
  specstyle/.style={
    colframe=darkgray, % Frame color
    coltitle=white, % Title text color
    fonttitle=\bfseries, % Title font
    boxrule=0.5mm, % Border thickness
    rounded corners,
    sharp corners=south, % Sharp bottom corners (optional)
    width=\linewidth, % Box width
    boxsep=4pt, % Box padding
    left=8pt, % Left padding
    right=8pt, % Right padding
    top=2pt, % Top padding
    bottom=2pt, % Bottom padding
    before=\bigskip, % Vertical space before the box
    after=\bigskip, % Vertical space after the box
    breakable,
  }
}

\newcommand\jsonnumber{\color{brown}}

\makeatletter
\newif\ifisvalue@json
\newif\ifisattr@json

\lstdefinelanguage{json}{
    tabsize             = 4,
    showstringspaces    = false,
    keywords            = {false,true},
    alsoletter          = 0123456789.,
    moredelim           = [is][\color{purple}]{`}{`},
    moredelim           = [is][\color{brown}]{``}{``},
    moredelim           = [is][\color{purple}]{[}{]},
    moredelim           = [is][\color{teal}]{"}{"},
    literate            = {,}{{,}}1,
    basicstyle          = \footnotesize\ttfamily,
    breaklines=true,
    breakatwhitespace=true
}

\newcommand\enterItemMode@json{%
    \leftbracket@json%
    \ifnum\lst@mode=\lst@Pmode%
        \global\isvalue@jsontrue%
    \fi
}

\newcommand\exitItemMode@json[1]{#1\global\isvalue@jsonfalse}

\newcommand\enterAttrMode@json{%
    \leftbrace@json%
    \ifnum\lst@mode=\lst@Pmode%
        \global\isattr@jsontrue%
    \fi
}

\newcommand\exitAttrMode@json[1]{#1\global\isattr@jsonfalse}

\lst@AddToHook{Output}{%
    \ifisvalue@json%
        \ifnum\lst@mode=\lst@Pmode%
            \def\lst@thestyle{\jsonnumber}%
        \fi
    \fi
    \lsthk@DetectKeywords% 
}

\makeatother

\lstdefinestyle{Prompt}{
  basicstyle        = \small\ttfamily,
  showstringspaces  = false,
  breaklines        = true,
  breakindent       = 0pt,
  rulecolor         = \color{black},
  escapeinside      = {(*@}{@*)},
  moredelim         = [is][\color{purple}\bfseries]{@@}{@@},
  moredelim         = [is][\color{teal}\bfseries]{(@}{@)},
  moredelim         = [is][\bfseries]{**}{**},
  moredelim         = [is][\color{red}\itshape]{|}{|},
}

\newtcolorbox[auto counter]{specbox}[2][]{%
    title=\graphshort{} Schema~\thetcbcounter: #2, #1}

\newtcolorbox[auto counter]{llmpromptbox}[2][]{%
    title=LLM Prompt ~\thetcbcounter: #2, #1}

\newtcolorbox[auto counter]{vqapromptbox}[2][]{%
    title=VQA Prompt ~\thetcbcounter: #2, #1}
\begin{document}

\runninghead{Loo et al.}
\title{\graphlong{}s for \\Open-World Object-Goal Navigation}
\author{Joel Loo$^{*}$\affilnum{1}, Zhanxin Wu$^{*}$\affilnum{2} and David Hsu\affilnum{1}}
\affiliation{\affilnum{1} Smart Systems Institute, National University of Singapore\\
\affilnum{2} College of Computing and Information Science, Cornell University}

\corrauth{David Hsu, Smart Systems Institute, National University of Singapore.}

\email{dyhsu@comp.nus.edu.sg}

\begin{abstract}

How can we build general-purpose robot systems for open-world semantic navigation, \eg, searching a novel environment for a target object specified in natural language? To tackle this challenge, we introduce \emph{\navsys{}}, a modular system composed of foundation models, for open-world Object-Goal Navigation (ObjectNav). Foundation models provide enormous semantic knowledge about the world, but struggle to organise and maintain spatial information effectively at scale. Key to \navsys{} is the \emph{\graphlong{}} representation, which acts as spatial memory for \navsys{}. It organises spatial information hierarchically using \emph{\graphshort{} schemas}, which are templates,  each describing the common structure of a class of environments. \graphshort{} schemas can be automatically generated from simple semantic labels of a given environment, \eg{}, ``home'' or ``supermarket''. They enable \navsys{} to adapt zero-shot to new environment types. We conducted experiments using both Fetch and Spot robots in simulation and in the real world, showing that \navsys{} achieves state-of-the-art performance on ObjectNav benchmarks and generalises zero-shot over diverse goals, environments, and robot embodiments.

\end{abstract}


\maketitle

\def\thefootnote{*}\footnotetext{Equal contribution.}

\section{Introduction}
\label{sec:intro}

\begin{figure}
  \includegraphics[width=\linewidth]{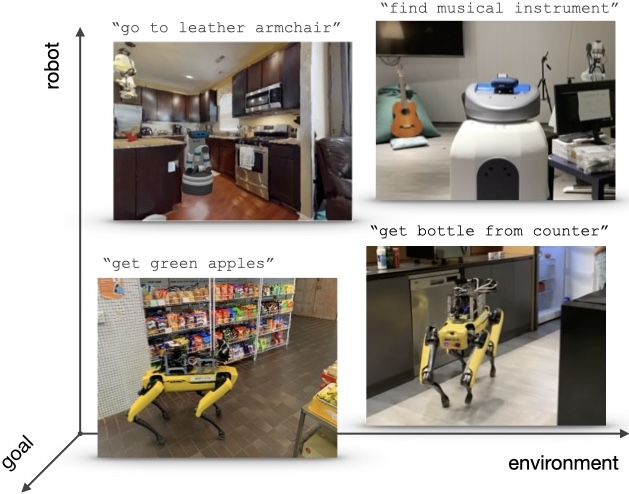}
  \caption{\textbf{ObjectNav in the open world.} Our experiments cover different combinations of goals, environments, and robots across both real world and simulation. We conducted experiments using both the Fetch and Spot robots in environments ranging from simulated homes (top-left) to real-world lab environments mimicking homes (top- and bottom-right) and stores (bottom-left).}
  \label{fig:splash_gen}
\end{figure}

Imagine a robot capable of taking verbal instructions from humans and quickly finding fever medicine in an unfamiliar pharmacy, finding a pack of cornflour in a new supermarket, or finding a bottle of wine in a friend's home---all without a prior map of the environment. To do this, the robot must understand the rich semantics of the world, in addition to its geometry.  This work focuses on \emph{object-goal navigation} (ObjectNav), a semantic navigation task, in the \emph{open world}. In ObjectNav, a robot navigates a \emph{novel} environment to find a  target object specified in  natural language. We extend the standard ObjectNav task~\citep{habitatchallenge2022, anderson2018evaluation} to the open-world setting with new challenges: the robot navigation system now needs to generalise over diverse goals, environments, and even robot embodiments (\autoref{fig:splash_gen}):

\begin{itemize}
    \setlength\itemsep{0pt}
    \item \emph{Goals}. The robot navigates to potentially any object described in natural language, including novel ones not known \textit{a priori}. 
    \item \textit{Environments}. The robot operates effectively in structurally different indoor environments---homes, offices, supermarkets, \etc{}---without environment-specific manual engineering.
    \item \textit{Robots}. The same navigation system works on different robot embodiments, wheeled or legged. 
\end{itemize}

\begin{figure*}
    \includegraphics[width=\linewidth]{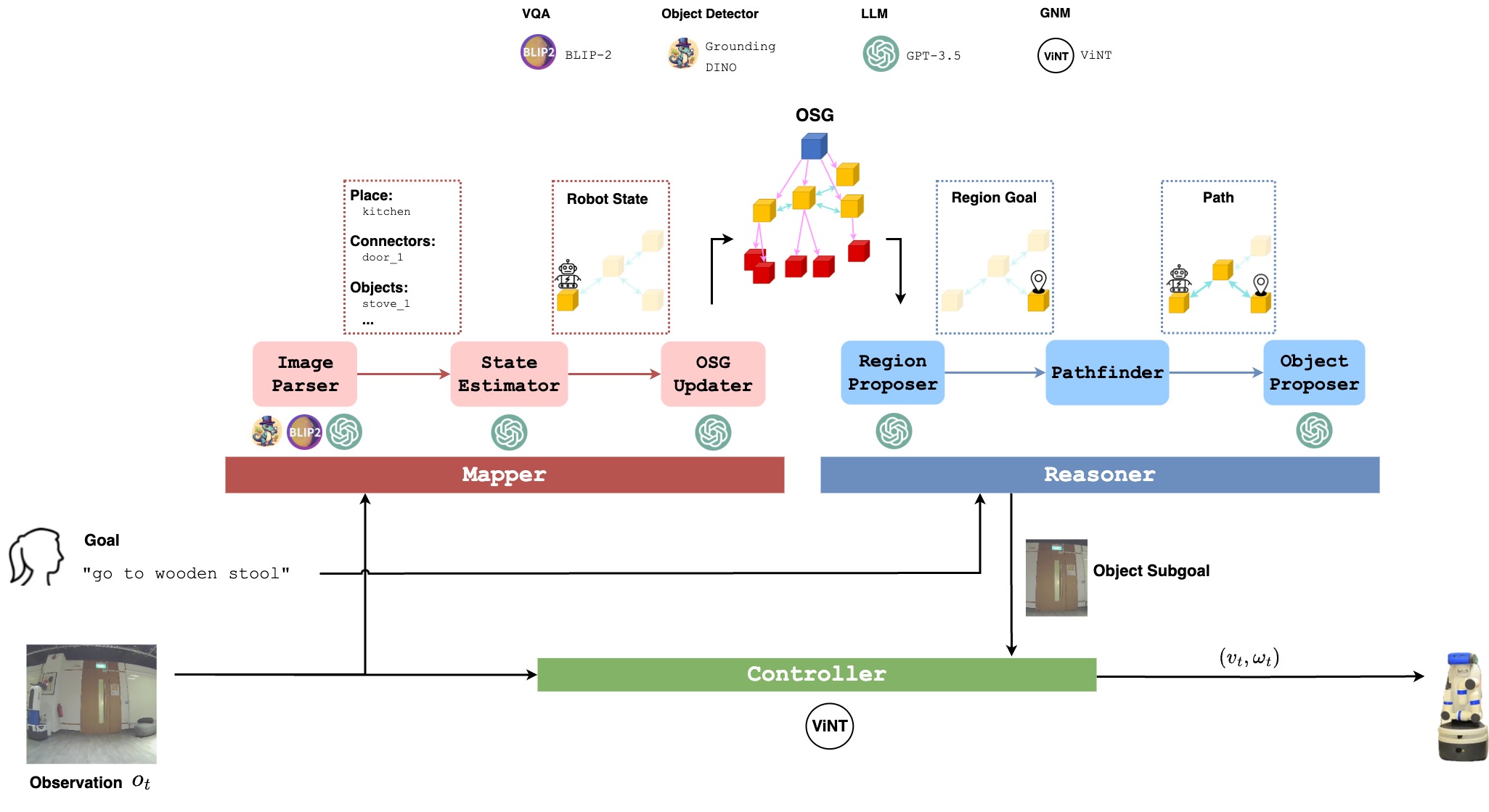}
    \caption{\textbf{ \navsys{}}. \navsys{} adopts a modular architecture, consisting of  modules built from foundation models: VFMs, LLMs, and GNMs.}
    \label{fig:system}
  \end{figure*}

%% OSG Navigator and foundation models
To this end, we introduce \emph{\navsys{}}, a modular navigation system composed of foundation models for robot perception, state estimation, planning, and local navigation control (\autoref{fig:system}).  Foundation models offer a promising approach: \textit{large language models} (LLMs) capture enormous semantic knowledge about the world~\citep{zhou2023navgpt, chen2024mapgpt, yu2023l3mvn, shah2023lfg},  while \textit{vision foundation models} (VFMs)~\citep{kirillov2023segany, liu2023grounding, li2023blip2, minderer2022owlvit} and \textit{general navigation models} (GNMs)~\citep{shah2023vint, sridhar2023nomad} show strong performance in robot perception and action across diverse environments and embodiments, respectively.  Specifically, \navsys{} uses two  VFMs, GroundingDINO~\citep{liu2023grounding} and BLIP-2~\citep{li2023blip2}, to process perceptual information; it uses an LLM, GPT-3.5~\citep{openai2024gpt4technicalreport}, to organise and reason about spatial information; it uses a GNM, ViNT~\citep{shah2023vint}, for low-level navigation control of  robots. Finally, it composes these models in a neuro-symbolic architecture for open-world ObjectNav. 

\begin{figure*}[!t]
  \includegraphics[width=\linewidth]{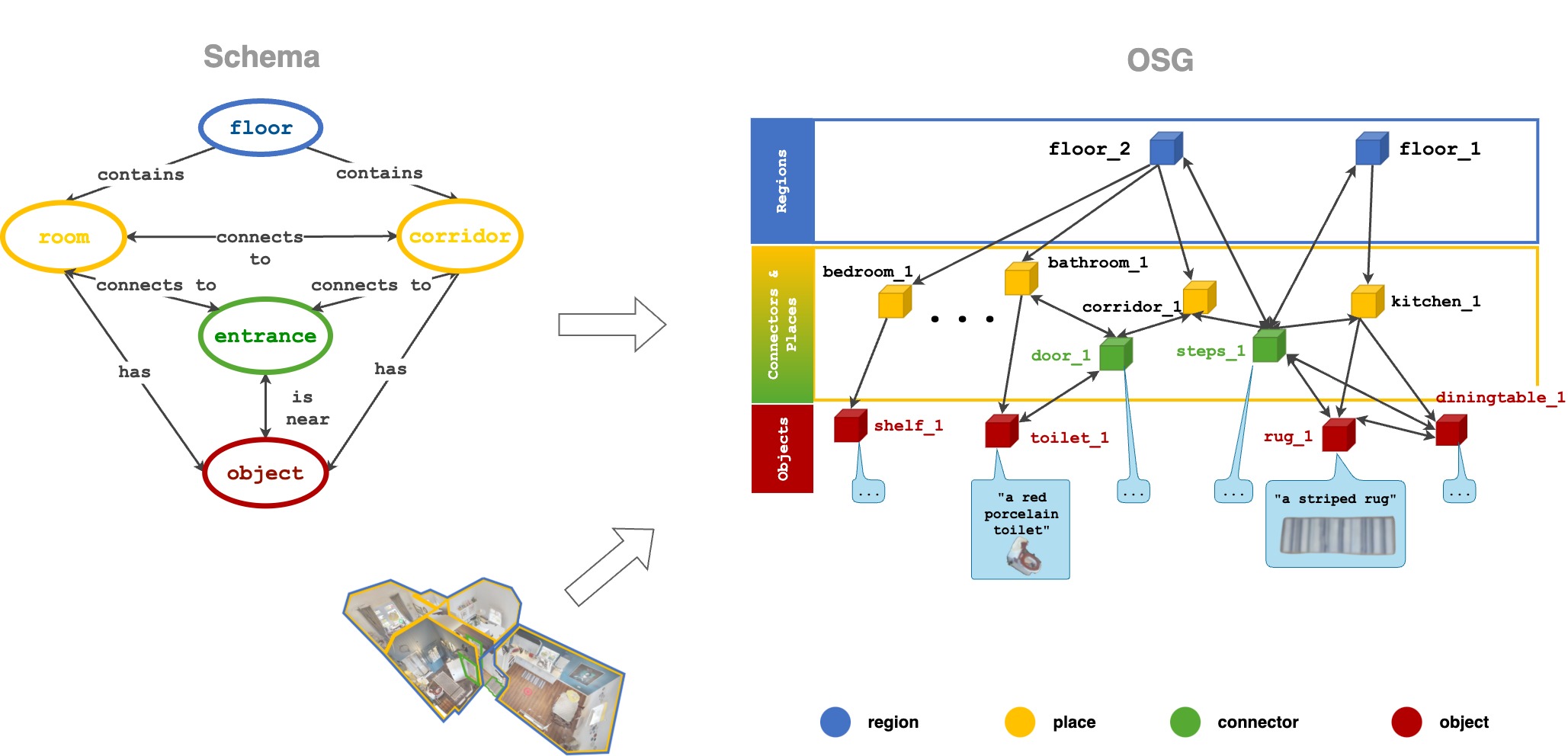}
  \caption{\textbf{An example \graphshort{} and its  schema.} Given a schema, \navsys{} constructs an OSG incrementally from visual observations of the environment. An OSG schema captures the common structure among environments within a class, \eg, home. }
  \label{fig:scene_graph_framework}
\end{figure*}

%% OSG
Foundation models are, however,  insufficient by themselves, as they struggle to organise and maintain spatial information effectively at scale. A key element of \navsys{} is the \textit{\graphlong{}} (\graphshort{}) representation (\autoref{fig:scene_graph_framework}).  A scene graph represents an environment as a hierarchically organised graph, in which  nodes represent semantic entities such as objects, places, regions, \etc{}, and  edges represent relations among them. Scene graphs serve as the unified memory coordinating the various foundation models in  \navsys{}, by maintaining a coherent spatial understanding of the environment. 

Traditional scene graphs are designed for specific environment types~\citep{armeni2019scenegraph, rosinol2021kimera, werby2024hovsg}. 
Each type has its own built-in semantic concepts: homes contain rooms and corridors, while supermarkets contain sections and aisles. This environment-specific design is impractical in the open world, as robots may encounter new environment types unknown in advance. To tackle this challenge, we introduce \textit{\graphshort{} schemas}, which are templates   capturing the common structure  among environments within a class~(\autoref{fig:scene_graph_framework}).  While schemas can be defined manually using domain-specific knowledge, we can again leverage foundation models, specifically, an LLM  to generate schemas automatically, given simple semantic labels of indoor scenes, \eg, ``home'' or ``supermarket''. This enables \navsys{} to adapt zero-shot to new types of  environments.

%% Contributions 
Our key contribution is \navsys{}, a neuro-symbolic system that combines foundation models and the OSG representation for open-world ObjectNav. \navsys{} adopts a modular architecture, consisting entirely of foundation models as components, for semantic navigation in novel environments without prior maps. It leverages \graphshort{} and the associated schema for flexible, structured environment representation. We evaluated \navsys{} in simulation and on two real robots, Fetch and Boston Dynamics Spot. Experimental results show  that it achieves state-of-the-art performance on ObjectNav benchmarks and generalises zero-shot over diverse goals, environments, and robot embodiments. Further analyses  point to the \graphshort{} representation as a key enabler of \navsys{}'s strong performance. The experiments suggest the following  main lessons:

\begin{itemize}
    \setlength\itemsep{0pt}
    \item Integrating  foundation models into the existing modular robot system architecture enables robot navigation that generalises zero-shot  across diverse environments (\autoref{sec:exp_objnav_results}), embodiments (\autoref{sec:exp_gen_embodiment}), and open-vocabulary goals (\autoref{sec:exp_gen_goals}).

    \item Foundation models, in particular, LLMs, benefit greatly from structured representations, such as \graphshort{}, in spatial planning (\autoref{sec:exp_objnav_results}).

    \item Semantics offers a promising approach to generalisable robot navigation, independent of metric geometry (Sections~\ref{sec:exp_objnav_results}--\ref{sec:exp_gen_embodiment}, \ref{sec:exp_data_assoc}, and \ref{sec:exp_realrobots}). 
    
\end{itemize}

Several limitations currently remain in \navsys{}. First, LLMs have made significant concurrent progress, and our current approach has yet to integrate the state of the art in multimodal models, multilingual models, or small, local models. Second, our approach has only been evaluated on indoor navigation tasks. Extending  \navsys{} to handle outdoor environments requires further investigation. Third, \navsys{} adopts a topo-semantic approach and does not capture metric information and geometry explicitly. Metric geometry is complementary and will enrich the \graphshort{} representation and further improve navigation performance. In our view, this last one offers the most exciting research opportunity in the immediate future. See \autoref{sec:conclusion} for further discussion.

In the following, \autoref{sec:related_work} reviews related work. \autoref{sec:approach} provides an overview of OSG Navigator. \autoref{sec:approach_osg} describes the OSG representation in details. Sections~\ref{sec:approach_osg_mapper} and \ref{sec:reason_control} present the main modules of OSG Navigator. \autoref{sec:result} presents the experimental results on open-world ObjectNav  and evaluates the effectiveness of the \graphshort{} representation. \autoref{sec:conclusion} summarizes the main findings and points out directions for future research.

\section{Related Work}
\label{sec:related_work}

\subsection{Semantic Navigation}
Robot navigation has expanded beyond tasks requiring geometric scene understanding and motion planning~\citep{hutchinson1996servo} to encompass richer, more complex \textit{semantic navigation} tasks~\citep{yokoyama2024vlfm}. These often specify abstract semantic goals, and require semantic reasoning and common-sense priors about environments to complete. Seminal examples include vision-and-language navigation~\citep{gu2022vln}, ImageNav~\citep{krantz2022instancespecific}, and ObjectNav~\citep{habitatchallenge2022} which respectively require robots to follow natural language instructions, reach a goal specified by an image, and find instances of a given object category. Later works build on these, increasing their complexity, and need for semantic information and reasoning by extending them to zero-shot~\citep{majumdar2022zson}, open-vocabulary~\citep{gadre2022cow}, multi-goal~\citep{wani2020multion}, and interactive~\citep{majumdar2023findthis} settings. We show that a robot system composed from foundation models and \graphlong{}s enables \textit{semantic navigation} in \textit{open world settings}, and demonstrate this on the task of open world ObjectNav (\autoref{fig:splash_gen}).

\subsection{Object-Goal Navigation}

ObjectNav~\citep{habitatchallenge2022} is the task of searching for an instance of a specified object category in a novel unmapped indoor environment. The skill of moving to a specified object is a useful and general primitive relevant to many downstream tasks in navigation and mobile manipulation~\citep{chang2023goat}. As such, ObjectNav has drawn an extensive body of work~\citep{wahid2018sac, mousavian2019visualrep, maksymets2021thda, ye2021auxiliary, ramakrishnan2022poni}, with notable progress made on benchmarks over the years. However, as \citet{majumdar2022zson} notes, these approaches typically require specific training on curated datasets for ObjectNav, limiting their ability to generalise to new scenes or goals. More broadly, the ObjectNav task is itself not fully \textit{open-world}, as it predefines a small set of object categories as goals, and does not consider generalisation across embodiments.

Foundation models' impressive zero-shot generalisation and open-world semantic knowledge~\citep{hu2023robotfmsurvey} are driving extensions of the ObjectNav task to open-world scenarios, and enabling the design of generalisable ObjectNav agents. \citet{majumdar2022zson, alhalah2022zsel} propose zero-shot task variants like Zero-shot ObjectNav (ZSON), which expands the set of goals to open-vocabulary object categories and requires that agents not rely on ObjectNav-specific demonstrations or rewards. \citet{gadre2022cow} extend this to more complex natural language queries in Language-driven ZSON (L-ZSON). Building on these, recent works explore designing agents to handle open-vocabulary object queries using LLMs~\citep{shah2023lfg, dorbala2024lgx, yu2023l3mvn} or VLMs~\citep{majumdar2022zson, khandelwal2022embclip, gadre2022cow, yokoyama2024vlfm} like CLIP~\citep{radford2021clip}. Our work goes further to address all three key dimensions of generalisation in the open world: diverse goals, environments and embodiments.

\subsection{LLMs in Navigation}

LLMs' open-world semantic knowledge and reasoning capabilities are valuable for semantic navigation~\citep{hu2023robotfmsurvey, huang2022survey}. They are a queryable source of semantic priors, for instance, offering priors on object co-occurrence in different environments~\citep{chen2023semutil, strader2024ontology, dorbala2024lgx, shah2023lfg}. They are also increasingly used to handle abstract, semantic concepts and plan with them. For instance, LLMs can be queried to generate plans satisfying abstract language instructions~\citep{ichter2022saycan, huang2022llmplanner, singh2023progprompt}, or to reason semantically about scene structure in search or exploration~\citep{zhou2023navgpt, chen2024mapgpt}. Recent works show that LLMs can be integrated with formal methods for planning~\citep{zhao2023llmmcts, huang2025limitlanguagemodelsplanning}, and for uncertainty handling~\citep{wang2025conformal} for more robust, structured reasoning. These capabilities make LLMs indispensable for semantic navigation tasks like ObjectNav, which require both strong semantic priors and planning.

How scene information is presented is a critical design choice in eliciting reasoning from LLMs~\citep{liu2022goodicl, strader2024ontology}. To overcome LLMs' limited context lengths, an external memory is needed to temporally accumulate scene information. While some approaches prompt LLMs with unordered lists of objects~\citep{cai2023pixnav, yu2023l3mvn, shah2023lfg, dorbala2024lgx}, incorporating spatial and semantic structure into the representation further enhances reasoning and reduces LLMs' hallucinations~\citep{rana2023sayplan, honerkamp2024momallm}. Thus, a structured open-set scene representation is needed for semantic navigation with LLMs. Recent works apply LLMs to pre-built scene graphs for efficient semantic search~\citep{rana2023sayplan}, or to translate ambiguous semantic tasks into formal grounded task specifications~\citep{dai2024optimal}. \citet{ravichandran2025spine, honerkamp2024momallm, rajvanshi2023saynav} apply LLMs to reason over graphical representations that are incrementally constructed, for under-specified, semantic navigation tasks.

\navsys{} achieves open-world ObjectNav by using LLMs to both construct generalisable, structured \graphshort{} representations, and to reason with \graphshort{}s. This work focuses on designing the representation and associated mapping algorithm. We leave improvements in reasoning to future work, and note that ongoing advances in integrating LLMs with planning and formal techniques can complement our work.

\subsection{Open-Set Scene Representations}
\label{sec:related_sg}

Open-set scene representations are commonly geometric maps with open-set semantic features fused into them~\citep{mascaro2024scenerepsurvey}. These include task-specific 2D costmaps~\citep{yokoyama2024vlfm, long2025instructnav, ren2024explore}, or 3D representations integrated with semantic embeddings---like point clouds~\citep{jatavallabhula2023conceptfusion, peng2023openscene}, NeRFs~\citep{shen2023F3RM}, and Gaussian splats~\citep{wu2021scenegraphfusion, qin2024langsplat}. However, LLMs struggle to parse geometric maps and are better suited at understanding semantics. Retrieval-augmented generation~\citep{lewis2020rag} offers vector databases as an alternative LLM-friendly representation~\citep{anwar2024remembr}. Embodied-RAG~\citep{xie2024embodiedrag} builds on this with a hierarchy of LLM-generated summaries for efficient indexing, though the lack of a schema to structure the hierarchy can lead to inconsistent abstractions.

Scene graphs are topological representations organising scenes in a hierarchy of spatial abstractions. As such they are suited for hierarchical reasoning for planning~\citep{mascaro2024scenerepsurvey, amiri2022reasoningscenegraphs, ravichandran2022hierarchical, honerkamp2024momallm, ray2024tampscenegraphs}. In particular, \citet{rana2023sayplan, rajvanshi2023saynav} demonstrate that scene graphs' symbolic and semantic nature facilitate task planning with LLMs. Recent progress has led to scene graphs that can be built incrementally~\citep{wu2021scenegraphfusion, hughes2022hydra}, and that capture open-set objects~\citep{gu2023conceptgraphs, werby2024hovsg}. However, a key weakness of most approaches, including Kimera~\citep{rosinol2021kimera} and HOVSG~\citep{werby2024hovsg}, is that they embed fixed schemas that only describe a specific environment class---often home/office environments comprised of rooms and floors. Recent works take steps toward enabling flexible schemas for mapping. \citet{strader2024ontology} proposes a learnable method to infer scene graph node labels that can be trained offline using only LLM guidance. Other approaches cluster geometric primitives based on a human-provided schema specifying open-vocabulary cluster labels---Lexis~\citep{kassab2024lexis} clusters rooms given a list of room labels, while Clio~\citep{maggio2024clio} clusters objects or location nodes based on a list of tasks.

\textit{\graphlong{}s} offer an open-world, structured representation in terms of consistent spatial abstractions. This is enabled by \graphshort{} schemas which, unlike Lexis and Clio, support complex schemas capturing \textit{hierarchies} of spatial abstractions.

\section{System Overview}
\label{sec:approach}
ObjectNav~\citep{yadav2022habitat} is the task of searching for a specified object category in novel unmapped indoor scenes. The inputs to an ObjectNav agent are RGB, depth, and noiseless positioning, while the outputs are velocity commands. We extend this to the open-world setting, requiring the agent to perform ObjectNav zero-shot across diverse environments, embodiments and goals. Our more challenging task setting allows only for RGB input, and linear/angular velocity command outputs, which are common denominators across a diverse range of robots. Our task setting is more realistic, as we do not assume access to privileged information like ground-truth localisation. The agent should also accept open-set object queries in natural language.

We formalise the task as an object-and-region-centric POMDP $\langle \oostate{}, \ooaction{}, \ootrans{} , \ooreward{}, \ooobs{}, \ooobsmodel{} \rangle$. This extends object-oriented POMDPs (OO-POMDPs)~\citep{wandzel2019mos, zheng2020mos, zheng2023genmos} to incorporate regions and relations which provide context about scene structure valuable for object search~\citep{grinvald2019volumetric, chen2023semutil}. The state comprises agent state ($s_a$), spatial regions ($s_r$), objects ($s_o$), and inter-object/region relations ($s_e$). Observations are RGB images, and actions include \textsc{MoveToObject}($o$) and \textsc{MoveToRegion}($r$). Transitions, $\ootrans{}$, are stochastic, and the observation model, $\ooobsmodel{}$, reflects uncertainty about the target object's location in novel, partially observable scenes. The goal, $\ooreward{}$, is an open-vocabulary object description.

Solving this POMDP by maintaining a belief over arbitrary scene structures and open-vocabulary goals is intractable. Instead, we forgo explicit belief tracking and planning, and leverage foundation models to update a scene memory that serves as a maximum likelihood estimate of the state, $\oostate{}$. An LLM-based policy selects actions based on this memory.

\textit{\navsys{}} instantiates this approach (\autoref{fig:system}) as a high-level mapping and reasoning layer coupled with a low-level navigation layer that implements the action primitives. To achieve the required generalisation, \navsys{} is modularly composed from foundation models, specifically four types of models: \textbf{(i)} LLMs, \textbf{(ii)} GNMs, \textbf{(iii)} VFMs for Visual Question Answering (VQA), and \textbf{(iv)} VFMs for Open-set Object Detection.

The high-level layer builds and reasons over \graphshort{}s, which serve as a generalisable, structured scene memory, and generates image subgoals which guide the low-level navigation layer. It does so with the \textit{\enginelong{}} and \textit{Reasoner} modules, which are templated algorithms that use \graphshort{} schemas to interface with the \graphshort{}. The \textit{Mapper} extracts open-vocabulary text describing salient scene elements from images, then uses LLMs to update the \graphshort{}. The \textit{Reasoner} uses LLMs to identify promising subgoals in the \graphshort{}, and plans a path over the \graphshort{} to reach it. The path is a sequence of image goals, with each goal an image crop of an object from an intermediate region along the path. The low-level navigation layer uses image goal-conditioned visuomotor controllers to drive the robot toward the subgoals. Specifically, it uses GNMs that execute at 10Hz. VLM perception in the higher layer operates at 2-2.5Hz, while LLM-based reasoning takes several seconds to update the scene graph and plan after reaching each subgoal, enabling near real-time object search.

\section{Open Scene Graphs}
\label{sec:approach_osg}

An \textit{\graphlong{}} (\graphshort{}) is a structured representation of open-set scene objects and spatial regions across multiple levels of abstraction. We define an ontology that abstractly describes the generic structure of indoor environments in terms of broad concept classes (\eg{} Objects, Places, Connectors, Region Abstractions) and relationship types (\eg{} proximity, connectivity, enclosure and hierarchy). This ontology provides the organisation principles for an \graphshort{}, and provides a framework in terms of which we can design and express a mapping algorithm for constructing \graphshort{}s.

For effective semantic reasoning, the \graphshort{} needs to describe scenes using specific semantic concepts. For instance, specific and meaningful concepts in terms of which to describe a home are rooms, corridors and floors (\autoref{fig:scene_graph_framework}). The abstract concepts in the ontology are \textit{specialised} for a particular environment type, yielding \textit{\graphshort{} schemas} that specify the spatial concepts and relationships characteristic to that environment. \autoref{listing:sim_homes} provides an example for general household environments. \graphshort{} schemas can be manually defined by the user, enabling them to incorporate domain knowledge about the particular class of environments the robot will encounter. However, we also propose an automated pipeline that leverages LLMs to generate \graphshort{} schemas automatically given simple semantic labels of indoor scenes (\autoref{sec:approach_auto_spec_gen}).

We first introduce the \graphshort{} framework in this section. \graphshort{}s can be constructed incrementally with an executable mapping routine described in the next section (\autoref{sec:approach_osg_mapper}). This is instantiated with an \graphshort{} schema, from a templated mapping algorithm specified in terms of the ontology. Formally, the resulting \graphshort{} instance $\mathcal{G}$, is a heterogeneous, simple, directed graph organized into $N$ layers, capturing a single scene from an environment type $E$ (\eg{} homes, supermarkets, hospitals).

\begin{specbox}[specstyle, label=listing:sim_homes]{Example schema for home environments}
\begin{lstlisting}[language=json]
`Floor`:
    ``layer_type``: "Region"
    ``layer_id``: 3
    ``contains``: [Room, Corridor]
    ``connects_to``: [Stairs]
`Room`:
    ``layer_type``: "Place"
    ``layer_id``: 2
    ``has``: [Object]
    ``connects_to``: [Entrance, Room, Stairs]
`Corridor`: 
    ``layer_type``: "Place"
    ``layer_id``: 2
    ``has``: [Object]
    ``connects_to``: [Entrance, Room, Stairs]
`Stairs`: 
    ``layer_type``: "Place"
    ``layer_id``: 2
    ``is_near``: [Object]
    ``connects_to``: [Floor, Room]
`Entrance`:
    ``layer_type``: "Connector"
    ``layer_id``: 2
    ``is_near``: [Object]
    ``connects_to``: [Room, Corridor]
`Object`:
    ``layer_id``: 1
\end{lstlisting}
\end{specbox}

\subsection{Abstract Node and Edge Types}
\label{sec:osg_node_edge_types}

\graphshort{}s represent information about \textit{objects} and \textit{locations}, where locations refer to sets of points that are fixed in space. We define four abstract node types in an \graphshort{}, each of which encapsulates different types and combinations of object and location information.

\begin{itemize}
    \item \textbf{\textit{Objects}}: These are semantically meaningful scene elements occupying spatially localised regions. \autoref{fig:scene_graph_framework} highlights a bed, sink and dining table as examples of Objects. To a robot, Objects provide semantic cues for reasoning, and can serve as navigation goals or localisation landmarks. In particular, the collection of neighbouring Objects can serve as a distinctive feature for localisation. This leads us to use \textit{object features}, \ie{} aggregated lists of nearby Objects, for state estimation and data association.

    \item \textbf{\textit{Places}}: These represent locations, specifically the most granular semantically meaningful spatial regions of a scene. Thus each scene can be spatially partitioned into a network of Places. Concretely, Places in homes can be \textit{rooms} or \textit{corridors} (\autoref{listing:sim_homes}) which do not consistently have semantically significant subregions. In supermarkets, Places may instead be \textit{aisles}. 

    \item \textbf{\textit{Connectors}}: These are localised structural elements that connect adjacent spatial regions and are therefore fixed on their boundaries, giving them properties of both objects and locations. Examples include \textit{entrances}, \textit{doors} and \textit{gates}. Connectors can serve as specific waypoints in space, and allow paths between regions to be finely described. Also, Connectors are also semantically meaningful and significant frontiers useful for exploration.

    \item \textbf{\textit{Region Abstractions}}: These represent coarse locations, and are higher-level spatial abstractions which contain multiple smaller spatial regions. For instance, a \textit{floor} may contain many \textit{rooms}, while a supermarket \textit{section} comprises a cluster of \textit{aisles}. Hierarchical abstractions like these can aid in reasoning, especially with LLMs~\citep{zheng2024stepback}.
\end{itemize}

We define four abstract edge types to capture various relations among objects and locations in a scene, which are encoded as directed edges in \graphshort{}s.

\begin{itemize}
    \item \textbf{\textit{Proximity}}: \edgenear{} edges are \textit{object-object} relations that approximately represent spatial closeness between objects. Concretely, the endpoints of such edges may be either Object or Connector nodes. This relation is used to select nearby nodes to use in \textit{object feature} landmarks, enabling us to distinguish objects and locations by their context.

    \item \textbf{\textit{Connectivity}}: \edgeconnects{} edges are \textit{location-location} relations that represent reachability between locations. This is a fundamental relation to enable navigation, also captured in representations ranging from occupancy grids to topological graphs. The endpoints of such edges may be any node representing locations, \ie{} Places, Connectors or Region Abstractions.

    \item \textbf{\textit{Inclusion}}: \edgehas{} edges connect \textit{locations to objects}, in particular identifying that an Object node (destination) is spatially contained within a Place node (source).
    
    \item \textbf{\textit{Hierarchy}}: \edgecontains{} edges are \textit{location-location} relations representing spatial containment of a region within another region. For example, it might connect a \textit{floor} Region Abstraction node to a \textit{room} Place node. These edges serve to define a tree-like hierarchy of spatial abstractions on the \graphshort{}, where each higher level captures a coarser abstraction. Source nodes must be Region Abstraction nodes, and destination nodes must be the location nodes at the next lower level in the tree, which may be either a Region Abstraction or Place node.
    
\end{itemize}

% Describe form of the OSG
\subsection{Layers and Structure of an \graphshort{}}
\label{sec:osg_structure}

We describe the structural constraints and hierarchy of \graphshort{}s, and detail the attributes maintained in them. This ensures that \graphshort{}s contain the information necessary to support robot functions.

\textbf{Layer 1: \textit{Objects.}} This layer captures the distinct objects in the scene, along with proximity relations among them. Object nodes are leaf nodes of the \graphshort{}, and each maintain four attributes: \textbf{(i)} an open-vocabulary text label for the object, \objnodelabel{} (\eg{} \textit{coffeetable, armchair}); \textbf{(ii)} an open-vocabulary text description of the object's appearance, \objnodetext{} (\eg{} \textit{white wooden} for a \textit{coffeetable} node); \textbf{(iii)} a unique node ID, \objnodeid{}; \textbf{(iv)} an image crop of the object, \objnodeimage{}.

We define an Object node's object features as the aggregated text labels and appearance descriptions of neighbouring nodes connected via \edgenear{} edges. \textit{I.e.}, [(\objnodelabel{}, \objnodetext{})$_{0}$, ..., (\objnodelabel{}, \objnodetext{})$_{K-1}$] for a node with $K$ neighbours. These features are distinctive landmarks to distinguish Object nodes.

\textbf{Layer 2: \textit{Places and Connectors.}} This layer captures fine-grained spatial location information, and describes the connectivity among locations in a scene. Both Place and Connector nodes may be found in this layer. During navigation, the robot's state is specified as a Place node from this layer.

Each Place node \placenode{} contains the attributes: \textbf{(i)} its class as a text string drawn from the \graphshort{} schema, \placenodeclass{} (\eg{} rooms, aisles); \textbf{(ii)} an open-vocabulary text label describing the Place \placenodelabel{}, (\eg{} \textit{living room} for rooms, \textit{dairy aisle} for aisles); \textbf{(iii)} a unique node ID, \placenodeid{}. Outgoing \edgeconnects{} edges to Place/Connector nodes define the scene connectivity and topology. Outgoing \edgecontains{} edges identify the Objects that lie within the Place node. Formally a Place node's \textit{object features} are the aggregated text labels and appearance descriptions of Object nodes contained within it, \ie{} nodes connected by \edgecontains{} edges.

As Connectors share object properties, they inherit the attributes and relations of Object nodes. They are also considered leaf nodes, which do not contain Objects. However as locations, they may also be linked to other locations with \edgeconnects{} edges. We note that not all environments have Connector nodes (\eg{} open-plan environments).

\textbf{Layers 3-$N$ (Optional): \textit{Region Abstractions.}} Each layer defines a different spatial abstraction, with abstractions in higher layers being more general. Each layer partitions the scene in terms of its associated abstraction. Specifically, the abstractions in layer $(i+1)$ should be a meaningful partition over nodes in layer $i$: \eg{} \textit{floors} forming a partition over the \textit{rooms} in a multi-storey house, and \textit{sections} serving as a partition over \textit{aisles} in a supermarket context. 

Thus nodes in the $(i+1)$th layer can only have \edgecontains{} edges to one or more nodes in the $i$th layer. Each node may be joined to Connector nodes via \edgeconnects{} edges to capture spatial connectivity---\eg{} from \autoref{fig:scene_graph_framework}, \textit{floor} abstraction nodes are linked via \textit{steps} (a Connector node). Each node contains at least the attributes: \textbf{(i)} an open-vocabulary text label describing the region \absinodelabel{}; \textbf{(ii)} a unique node ID, \absinodeid{}.

\section{Topo-Semantic Mapping and Localisation}
\label{sec:approach_osg_mapper}

\begin{algorithm}
\caption{\textit{\engineshort{}}}
\label{alg:mapper}
\begin{algorithmic}[1]
    \REQUIRE Image $I_t$, \graphshort{} schema $\mathcal{S}$, previous \graphshort{} $\mathcal{G}$, previous robot state (Place) $s^a_{t-1}$
    \ENSURE Updated \graphshort{} $\mathcal{G}^{'}$, updated current state $s^a_t$ \\
    \COMMENT{Identify Objects, Connectors and Places}
    \STATE $\mathcal{O}_{parsed} \leftarrow$ \textsc{ImageParser}($\mathcal{S}, I_t$) \\
    \COMMENT{Estimate state (\ie{} current occupied Place node)}
    \STATE $\hat{s}^a_t\leftarrow$ \textsc{StateEstimator}($\mathcal{S}, \mathcal{G}, s^a_{t-1}, \mathcal{O}_{parsed}$) \\
    \COMMENT{Integrate observations into the \graphshort{}}
    \STATE $\mathcal{G}^{'}, s^a_t\leftarrow$ \textsc{OSGUpdater}($\mathcal{S}, \mathcal{G}, \hat{s}^a_t, s^a_{t-1}, \mathcal{O}_{parsed}$)
    \RETURN $\mathcal{G}^{'}, s^a_t$
\end{algorithmic}
\end{algorithm}

\navsys{} performs topo-semantic localisation and mapping, constructing and updating an \graphshort{} state representation online during navigation using the \textit{\enginelong{}}. The \engineshort{} (\autoref{alg:mapper}) routine given an \graphshort{} schema. The schema defines environment-specific spatial abstractions, filling in the template's parameters with concrete spatial concepts and their relational structure. Since these concepts are open-vocabulary, the \engineshort{} leverages LLMs to interpret and reason about them.

\autoref{sec:approach_auto_spec_gen} describes an offline \graphshort{} schema generation pipeline, that draws on LLMs' open-world knowledge of indoor environments to construct schemas given a text label of the environment type. This enables the \engineshort{}, and by extension the \navsys{}, to generalise zero-shot to new environments \textit{without human guidance}.

The \engineshort{} constructs \graphshort{} online, and comprises three components (\autoref{fig:system}): \textbf{(1)} an \textit{image parser} to extract semantic information from RGB observations (\autoref{sec:approach_image_parser}); \textbf{(2)} a \textit{state estimator} (\autoref{sec:approach_osg_mapper_state_est}); \textbf{(3)} an \textit{\graphshort{} updater} to integrate parsed information into the \graphshort{} (\autoref{sec:approach_osg_updater}). These components rely on foundation models---LLMs, VQA models and open-set object detection (dependencies highlighted in \textcolor{brown}{brown} in the algorithms below). While the \engineshort{} is deterministic, we propose an approach in \autoref{sec:approach_mapper_prob} to extend it to incorporate probabilities and maintain multiple hypotheses about the state and \graphshort{}.

\subsection{Automatic \graphshort{} Schema Generation}
\label{sec:approach_auto_spec_gen}

\begin{figure}[!t]
  \includegraphics[width=\linewidth]{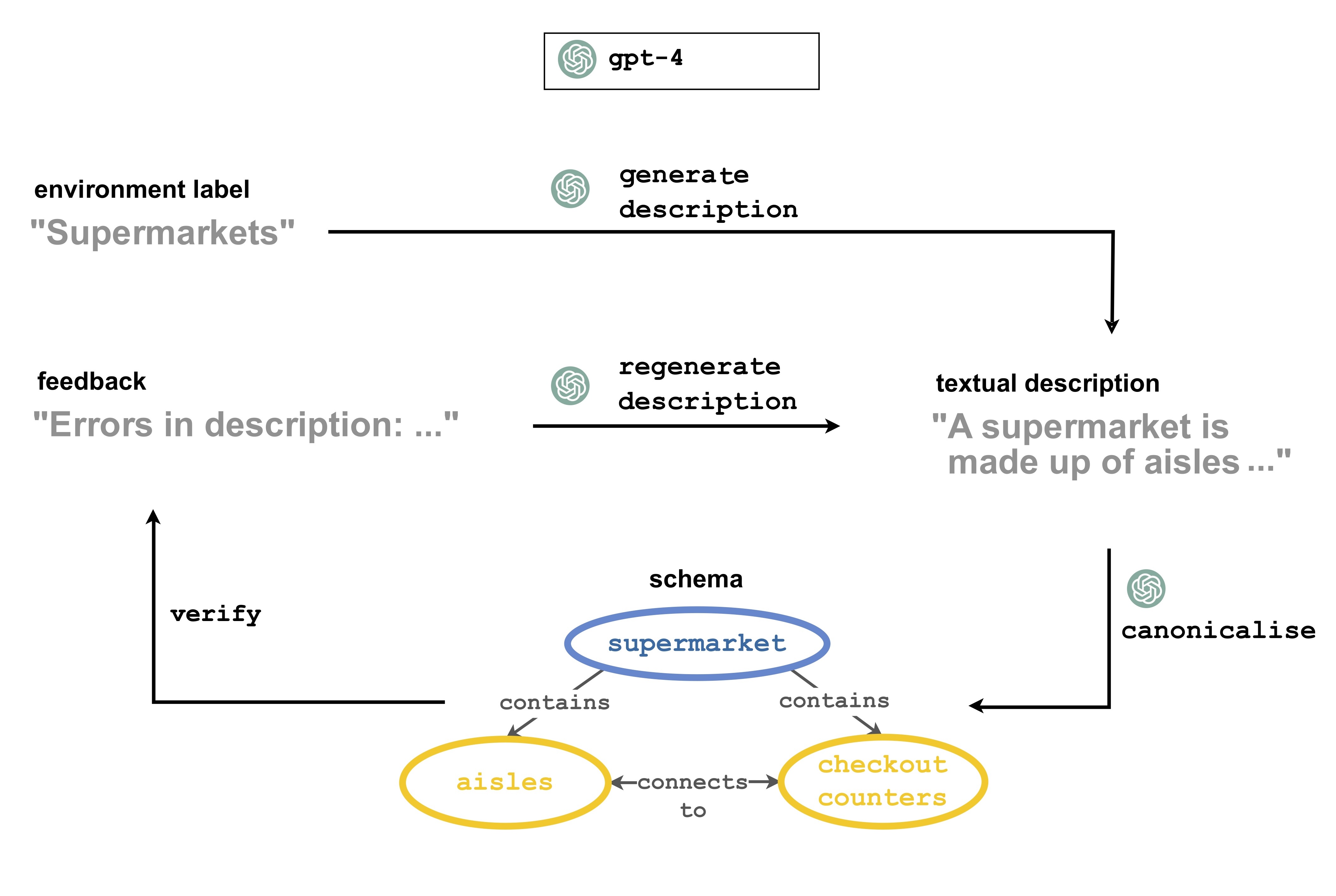}
  \caption{\textbf{ Automatic \graphshort{} schema generation.} \graphshort{} schemas are generated from environment labels, \eg{} ``supermarkets'', via an iterative process.
  }
 
  \label{fig:auto_spec_gen}
\end{figure}

The \engineshort{}'s ability to build a coherent and descriptive scene graph relies on having a sound and sufficiently representative \graphshort{} schema for that environment. Our \graphshort{} schema generator can automate the design process for such schemas: given only a simple text label describing the environment type, \eg{} ``homes'' or ``supermarkets'', the generator produces an \graphshort{} schema for that environment. Specifically, the generator produces a JSON-structured \graphshort{} schema that is \textit{correct}, \ie{} satisfies \graphshort{}'s structural requirements, while providing \textit{coverage}, \ie{} uses suitable spatial abstractions and provides sufficient detail to describe the typical structure of this environment type. Intuitively, we draw on LLMs' prior knowledge of environments and ability to organise open-vocabulary information for this task. Our approach (\autoref{fig:auto_spec_gen}) decomposes the generator into a three-stage pipeline, where each stage focuses on optimising either correctness or coverage. The pipeline is run iteratively to refine the generated schema until it complies with the \graphshort{}'s requirements. While the current pipeline requires an input label of the environment type, we note that this can be further automated by using VLMs to classify the environment based on sensor input.

\textbf{Stage 1: Description Generation.} We prompt the LLM to generate an unstructured, long-form natural language description of the given environment type (\autoref{listing:auto_gen_description_struct}). It is instructed to provide details on the typical spatial abstractions used to describe such environments, and possible relations between these abstractions. We hypothesise that this text generation task is more in-distribution for LLMs~\citep{brown2020fewshotlearners} and an effective way to query large amounts of common-sense knowledge.

\textbf{Stage 2: Canonicalise.} We guide the LLM to convert the unstructured description into an \graphshort{} schema in the form of a canonical graph representation that captures identified spatial abstractions as nodes and relations among them as edges. We take a similar approach to \citet{zhang2024extract}: the LLM first extracts triplets [\textit{Abstraction1}, \textit{Relation}, \textit{Abstraction2}] from the text (\autoref{listing:auto_gen_triplets_struct}), then canonicalises the abstractions and relations (\autoref{listing:auto_gen_canonicalisation_struct}). Specifically, the LLM maps spatial abstractions to specified node types (\textit{Objects}, \textit{Connectors}, \textit{Places}, \textit{Region Abstractions}), and selects the subset of relations corresponding to allowed \graphshort{} edge types (\eg{} \texttt{connects to}, \texttt{contains}). We assemble the canonical graph from the canonicalised triplets.

\textbf{Stage 3: Verify.} The canonical graph is checked against the \graphshort{}'s structural requirements. The \graphshort{} implements a verifier class to parse \graphshort{} schemas and provide natural language feedback for detected violations of an \graphshort{}'s requirements. If errors are detected, the feedback is appended to the textual description from Stage 1, and the LLM is prompted to re-generate the description to correct the errors. Otherwise, the \graphshort{} schema is returned in a JSON format.

\subsection{Image Parser}
\label{sec:approach_image_parser}
This extracts image and open-vocabulary textual information about Places, Objects and Connectors from RGB images. It relies on VFMs to parse visual inputs: an \textit{open-set object detector} --- GroundingDINO~\citep{oquab2023dinov2} --- localises and labels scene elements, while a \textit{VQA model} --- BLIP2~\citep{li2023blip2} --- translates image observations into compact language descriptions. We provide details on incorporating extracted images and textual information into \graphshort{} in \autoref{sec:approach_osg_mapper_state_est} and \autoref{sec:approach_osg_updater}.

\textbf{Parsing Place.} We obtain a semantically meaningful text label for the current Place node by querying a VQA model with RGB images of the robot's surroundings. As the \graphshort{} schema may contain multiple types of Places, \eg{} \textit{rooms}, \textit{corridors} \etc{}, we first ask the VQA model to select the Place type that best describes the observations (\autoref{listing:place_class_struct}). Given this inferred type, the VQA model is then queried for a more specific label of the currently occupied Place that reflects its function or semantic meaning (\autoref{listing:place_label_struct}). \textit{E.g.}, given that the current Place is a \textit{room} instance, the VQA is required to label it specifically, as a \textit{living-room} or \textit{kitchen} for example.

\textbf{Parsing Objects/Connectors.} An open-set object detector identifies and labels Objects and Connectors in the scene. As the detector does not distinguish between Objects and Connectors, we prompt an LLM to sort the detections into the two types of leaf nodes (\autoref{listing:classify_elements_struct}). The VQA model is queried to describe image crops of each Object/Connector, providing the textual attributes for each node (\autoref{listing:obj_description_struct}). We use a heuristic that connects leaf nodes with \edgenear{} edges if pixel distances between their corresponding bounding boxes fall below a threshold.

\textbf{Goal Detection.} The image parser prompts an LLM with the detected objects and their descriptions, and asks it to identify whether any object in the set matches the described goal well. The agent will navigate directly to any detected goal and terminate the episode. We find empirically that goals specifying commonplace object categories can robustly be detected using string matching on the object labels output by the BLIP-2 VQA model, to a similar success rate as LLMs.

\subsection{State Estimator} 
\label{sec:approach_osg_mapper_state_est}

\begin{algorithm}
\caption{\textit{State Estimator}}
\label{alg:mapper_state_est}
\begin{algorithmic}[1]
\REQUIRE \graphshort{} schema $\mathcal{S}$, \graphshort{} $\mathcal{G}$, previous robot state $s^a_{t-1}$, parsed observations $(P, O, C)$
\ENSURE Estimated robot state $\hat{s}^a_t$

\STATE $\mathcal{V}_P \leftarrow$  \textsc{GetPlacesLayer}($\mathcal{G}$) \\
\COMMENT{Object features from observations}
\STATE $F_{obs} \leftarrow O\cup C$\\
\COMMENT{Get semantically similar nodes (\autoref{listing:place_name_struct})}
\STATE $\mathcal{V}_{sim} \leftarrow$ \textcolor{brown}{\textsc{GetSimilarPlaces\_LLM}}($\mathcal{S}, P, \mathcal{V}_P$) \\
\COMMENT{Sort nodes by increasing distance from $s^a_{t-1}$}
\STATE $\mathcal{V}_{sorted} \leftarrow$ \textsc{SortByDistance}($\mathcal{V}_{sim}, s^a_{t-1}$)
\FOR{$v$ \textbf{in} $\mathcal{V}_{sorted}$:}
    \STATE $F_{obj} \leftarrow$ \textsc{GetObjectFeatures}($\mathcal{G}, v$) \\
    \COMMENT{Object feature matching (\autoref{listing:place_recognition_struct})}
    \IF{\textcolor{brown}{\textsc{PairwisePlaceMatch\_LLM}}($\mathcal{S}, F_{obj}, F_{obs}$):}
        \RETURN $v$
    \ENDIF
\ENDFOR
\RETURN \texttt{None}
\end{algorithmic}
\end{algorithm}

The state estimator compares textual observations extracted by the Image Parser from visual input, with the \graphshort{} to determine the robot's state $s_t^a$, which is the currently occupied Place node. Conceptually, the robot's state is identified through place recognition, based on matching textual observations of the surroundings with the \textit{object features} of possible state nodes in the \graphshort{}.

In practice, the state estimator only attempts to match with semantically similar Place nodes in the \graphshort{}, greedily prioritising those likely to be nearby (\autoref{alg:mapper_state_est}). LLMs play two key roles in state estimation, being employed in both cases as \textit{similarity metrics} for semantic concepts expressed in ambiguous natural language. Firstly, LLMs help narrow down the set of Place nodes to those with semantically similar labels to the Place label inferred from observations. For instance, if the robot observes that it is in a \textit{bedroom}, the LLM should select nodes of similar function or meaning like \textit{bedrooms} and \textit{guestrooms} as plausible robot states. Secondly, LLMs match Place nodes with observations in a pairwise fashion. They are guided to make a binary decision on the similarity between observed objects and each Place node's object features, with emphasis on larger objects.

\subsection{\graphshort{} Updater}
\label{sec:approach_osg_updater}

\begin{algorithm}
\caption{\textit{OSG Updater}}
\label{alg:mapper_osg_updater}
\begin{algorithmic}[1]
\REQUIRE \graphshort{} schema $\mathcal{S}$, \graphshort{} $\mathcal{G}$, estimated state $\hat{s}^a_t$, previous state $s^a_{t-1}$, parsed observations $(P, O, C)$
\ENSURE Updated current state $s^a_t$, updated \graphshort{} $\mathcal{G}^{'}$
\IF{$\hat{s}^a_t$ \textbf{is} \texttt{None}:} 
    \STATE \COMMENT{Unvisited area. Add new Place and leaf (Object and Connector) nodes.}
    \STATE $s^a_t \leftarrow \textsc{AddPlaceNode}(\mathcal{G}, P)$
    \STATE \textsc{AddLeafNodesAndEdges}($\mathcal{S}, \mathcal{G}, O\cup C$)
    \STATE \COMMENT{Update successively higher region layers, using LLMs to form abstractions (\autoref{listing:region_abs_struct})}
    \STATE $v_{t}, v_{t-1} \leftarrow s^a_t, s^a_{t-1}$
    \FOR{$i$ in \texttt{range}(4, $\textsc{GetNumLayers}(\mathcal{S})+1$):}
        \STATE $\mathcal{V}_i \leftarrow$ \textsc{GetLayer}($\mathcal{G}, i$)
        \STATE $v_{abs} \leftarrow$ \textcolor{brown}{\textsc{InferRegion\_LLM}}($\mathcal{S}, \mathcal{V}_i, v_{t}, v_{t-1}$)
        \IF{$v_{abs}$ \textbf{in} $\mathcal{V}_i$:}
            \STATE \COMMENT{$s^a_t$ in existing region abstraction node}
            \STATE \textsc{AddEdge}($\mathcal{S}, \mathcal{G}, v_{abs}, v_{t}$)
            \STATE \textbf{break}
        \ELSE
            \STATE \COMMENT{Add new region abstraction node}
            \STATE \textsc{AddNode}($\mathcal{V}_i, v_{abs}$)
            \STATE \textsc{AddEdge}($\mathcal{S}, \mathcal{G}, v_{abs}, v_t$) 
            \STATE \textsc{AddEdge}($\mathcal{S}, \mathcal{G}, v_{abs}, v_{t-1}$)
        \ENDIF
        \STATE $v_{t}, v_{t-1} \leftarrow v_{abs}, \textsc{GetParent}(\mathcal{G}, v_{t-1})$ 
    \ENDFOR
    
\ELSE
    \STATE \COMMENT{Previously visited Place. Associate observations with leaf nodes to update current Place's contents}
    \STATE $s^a_t\leftarrow \hat{s}^a_t$
    \STATE $\mathcal{V}_{\text{leaf}}, \mathcal{F}_{\text{leaf}} \leftarrow$ \textsc{GetLeafNodesAndFeats}($\mathcal{G}, s^a_t$)
    \FOR{$o$ \textbf{in} $O \cup C$:}
        \STATE $f_o \leftarrow$ \textsc{GetObjectFeatures}($o, O\cup C$)
        \STATE \COMMENT{Associate $o$ with leaf nodes (\autoref{listing:data_assoc_ex})}
        \STATE $o' \leftarrow$ \textcolor{brown}{\textsc{MatchObjects\_LLM}}($o, f_o, \mathcal{V}_{\text{leaf}}, \mathcal{F}_{\text{leaf}}$)
        \IF{$o'$ \textbf{is} \texttt{None}}
            \STATE \textsc{AddLeafNodesAndEdges}($\mathcal{S}, \mathcal{G}, \{o\}$)
        \ELSE
            \STATE $\mathcal{E}\leftarrow$ \textsc{GetObservedEdges}($\mathcal{G}, o, O\cup C$)
            \STATE \textsc{UpdateLeafNodesAndEdges}($\mathcal{G}, o', o, \mathcal{E}$)
        \ENDIF
    \ENDFOR
\ENDIF
\RETURN $\mathcal{G}, s^a_t$
\end{algorithmic}
\end{algorithm}

The \graphshort{} updater incrementally integrates observations from the Image Parser into the \graphshort{}. Conceptually, it first updates the \graphshort{} with directly observable scene elements---\ie{} Places, Objects, Connectors, and the edges between them---then abstracts them using the semantic spatial abstractions from the \graphshort{} schema (\autoref{alg:mapper_osg_updater}). Objects/Connectors and their spatial proximity relations, as well as Places, are directly updated from the parsed observations. Spatial connectivity between Places and Connectors is determined from the parsed observations, estimated state and history of previous robot states, following the valid connectivity rules from the \graphshort{} schema.

Two challenges underlie this approach: \textbf{(i)} data association, to identify if an observed scene element is seen and thus determine whether to add a new node or update an existing one; \textbf{(ii)} reasoning to describe and abstract away spatial regions using semantic abstractions. As both involve dealing with semantic information and concepts expressed in natural language, we address them with LLMs.

\textbf{Data Association.} We hypothesise that a scene element's \textit{object features}, \ie{} a list of nearby Objects/Connectors and their descriptions, serves as a distinctive signature by which they may be recognised. We use the LLM to gauge \textit{semantic similarity} between the object features of an observed Object/Connector and those of the leaf nodes in the robot's vicinity (\autoref{alg:mapper_osg_updater}, Line 27). The LLM is prompted to return the associated leaf node $v\in \mathcal{V}_{\text{leaf}}\cup\{\varnothing\}$, where $\mathcal{V}_{\text{leaf}}$ is the set of all leaf nodes.

\textbf{Abstracting Spatial Regions.} To construct the \graphshort{}'s hierarchy of spatial abstractions, we partition layer $i$'s nodes based on the abstraction specified for layer $(i+1)$ in the \graphshort{} schema. Spatial abstractions can sometimes be inferred from spatial connectivity and topology, based on the structure given in the \graphshort{} schema. For instance in \autoref{fig:scene_graph_framework}, the \textit{stairs} Connector serves as the marker to segment the environment into different \textit{floor} abstraction nodes. In other cases however, semantic reasoning is needed. For instance, the different \textit{sections} in a supermarket (\eg{} deli, household items, produce sections) may not have clear structures like Connectors to delimit their boundaries. Thus semantic reasoning may be required to cluster the individual \textit{aisles} into such semantically meaningful \textit{sections}. We prompt LLMs to perform this function in a greedy manner (\autoref{alg:mapper_osg_updater}, Line 9): for each new state the robot enters, the LLM decides if it belongs to an existing abstraction node, or if the current state should belong to a new abstraction node.

\subsection{Probabilistic \graphshort{} Mapping}
\label{sec:approach_mapper_prob}

Since our focus is on incorporating flexible semantics and structure into scene graphs, we design the \engineshort{} as a deterministic algorithm for simplicity. Practically, it is capable of building accurate scene graphs over diverse environments, partly due to LLMs' strong ability to handle ambiguity and observational uncertainty (\autoref{sec:exp_osg_building}). However, we also propose a means to incorporate uncertainty handling into the \engineshort{} for improved robustness, by extending it to consider perceptual errors and uncertainty. Specifically, we extend \graphshort{}s to maintain multiple hypotheses about the \textit{topology} of the scene.

We build on \textit{Online Probabilistic Topological Mapping} (OPTM)~\citep{ranganathan2011optm}, a framework which maintains multiple hypotheses about possible map topologies and computes a posterior over them, using a particle filter where each particle represents a specific map topology. Concretely, a map topology $\mathcal{T}$ is uniquely defined by a partition of the robot's observations accumulated over time, \ie{} $\mathcal{O} =\{o_1, o_2, ..., o_T\}$. \textit{E.g.}, 
\begin{equation}
    \mathcal{T}_T = \{v_1, ..., v_\xi\} = \{\{o_1, o_2, o_T\}, \{o_3, o_5\}, ...\}
\end{equation} 

Each $v_i$ represents a node in $\mathcal{T}_T$ and edges between nodes are implicitly represented through consecutive observations $o_{i-1}, o_i$ that belong to separate nodes.

Like standard particle filters, OPTM first performs a predict step, which predicts likely new map topologies $\mathcal{T}_t$ for each particle, based on their current topology $\mathcal{T}_{t-1}$. Next, it updates each particle's weight based on how well the observations describe the topology it represents, then resamples particles to focus on more probable topologies. Specifically in a probabilistic \graphshort{}, each topology comprising nodes from the Places and Connectors layer. Thus each particle represents a hypothesis on the fine-grained spatial connectivity of the scene. Observations $o_t$ correspond to the parsed output from the Image Parser.

\textbf{Prediction.} For each particle the filter predicts how its topology is likely to be updated at the next timestep $t$, by sampling from the proposal distribution $p(\mathcal{T}_t | \mathcal{T}_{t-1})$. This amounts to selecting the $v\in \mathcal{T}_{t-1}\cup\{\varnothing\} = \{v_1, ..., v_\xi, \varnothing\}$ to which observation $o_t$ belongs, where $v_i$ is an existing node and the empty set is a new node. The proposal distribution is designed to favour topologies where $o_t$ is added to frequently visited nodes (\ie{} the partitions $v_i$ with high $|v_i|$), and nodes in close proximity to the robot's previous state. The former captures the assumption that the robot visits fewer new areas over time and that highly visited nodes are central areas more likely to be revisited (\eg{} a corridor or lobby area), while the latter is an informal prior on the robot's possible range of motion over a short duration. Like OPTM, we model the former with a Dirichlet process, and model the latter with a uniform distribution over nearby nodes in the graph. This heuristic proposal can potentially be extended to include observations for targeted sampling, \eg{} increasing weight on nodes with the same Place label as the current observation's.

\textbf{Weight Update and Resampling.} Similar to OPTM, importance weights are updated as,
\begin{equation}
    w_t^{(i)} \propto p(o_t | \mathcal{T}^{(i)}_t)w_{t-1}^{(i)}\text{,  for the $i$th particle}
\end{equation}

Let node $v'\in\mathcal{T}_t$ be the current node to which $o_t$ belongs, and let $\mathcal{V} \subseteq \mathcal{T}_t$ be nodes with semantically similar labels to $o_t$'s Place label. We define the observation model as,
\begin{equation}
    p(o_t|\mathcal{T}^{(i)}_t) = p(o_t|v')\prod\limits_{v_i\in\mathcal{V}\setminus\{v'\}} (1 - p(o_t|v_i))
\end{equation}

This captures the notion that $o_t$ is most likely to have been generated from node $v'$, \ie{} it matches best the features observed at $v'$ while matching more poorly with other nodes. This can be viewed as a probabilistic extension to the greedy Place estimation algorithm in \autoref{sec:approach_osg_mapper_state_est}. Similar to it, we query LLMs to match observations to Place object features with Yes/No questions and obtain $p(o_t|v)$ as the normalised logits to its response. However, this extension differs in that it fully considers all nodes in $\mathcal{V}$ and accounts for their probabilities. The updated importance weights are then used to resample particles to focus on probable topologies.

\section{Reasoning and Control}
\label{sec:reason_control}

\subsection{Reasoner}
\label{sec:approach_reasoner}

\begin{algorithm}
\caption{\textit{Reasoner}}
\label{alg:reasoner}
\begin{algorithmic}[1]
    \REQUIRE \graphshort{} schema $\mathcal{S}$, \graphshort{} $\mathcal{G}$, current state $s^a_t$, target region $r_{\text{subgoal}}$, object goal $g$
    \ENSURE Object goal image $I_o$, target region to explore $r_{\text{subgoal}}$
    \IF{$r_{\text{subgoal}}$ \textbf{is} \texttt{None}:}
        \STATE \COMMENT{Identify promising region in $\mathcal{G}$ to search for goal $g$}
        \STATE $r_{\text{subgoal}} \leftarrow$ \textcolor{brown}{\textsc{RegionProposer\_LLM}}($\mathcal{S}, \mathcal{G}, g$) (\autoref{alg:region_proposer})
    \ENDIF
    \IF{\textsc{ReachedTargetRegion}($r_{\text{subgoal}}, s^a_t$):}
        \STATE \COMMENT{Query LLM for an object in $\mathcal{G}$ around which to search for $g$ (\autoref{listing:object_prompt_struct})}
        \STATE $I_o \leftarrow \textcolor{brown}{\textsc{ObjectProposer\_LLM}}(\mathcal{S}, \mathcal{G}, r_{\text{subgoal}})$
        \STATE $r_{\text{subgoal}} \leftarrow$ \texttt{None}
    \ELSE
        \STATE \COMMENT{Plan path to $r_{\text{subgoal}}$ and return next waypoint}
        \STATE $r_{\text{wp}} \leftarrow \textsc{Pathfinder}(\mathcal{G}, r_{\text{subgoal}}, s^a_t)$
        \STATE \COMMENT{Query LLM for an Object/Connector that takes the robot toward $r_{subgoal}$ (\autoref{listing:object_prompt_struct})}
        \STATE $I_o \leftarrow \textcolor{brown}{\textsc{ObjectProposer\_LLM}}(\mathcal{S}, \mathcal{G}, r_{\text{wp}})$
    \ENDIF    
    \RETURN $I_o, r_{\text{subgoal}}$
\end{algorithmic}
\end{algorithm}

\begin{algorithm}
\caption{\textcolor{brown}{\textsc{RegionProposer\_LLM}}}
\label{alg:region_proposer}
\begin{algorithmic}[1]
    \REQUIRE \graphshort{} schema $\mathcal{S}$, \graphshort{} $\mathcal{G}$, object goal $g$
    \ENSURE Target region to explore $r_{\text{subgoal}}$

    \STATE $N\leftarrow \textsc{GetNumLayers}(\mathcal{S})$
    \STATE \COMMENT{Initialise subgraph for search}
    \STATE $\mathcal{G}_{search} \leftarrow \textsc{GetLayer}(\mathcal{G}, N)$
    \STATE \COMMENT{Iterate over region layers, from highest to lowest}
    \FOR{$i$ \textbf{in} \texttt{range}(N, 1):}
        \STATE \COMMENT{Query LLM for most promising region to explore at current layer (\autoref{listing:region_prompt_struct})}
        \STATE $r_i \leftarrow \textcolor{brown}{\textsc{SelectRegion\_LLM}}(\mathcal{G}_{search}, g)$
        \STATE $\mathcal{G}_{search} \leftarrow \textsc{GetChildNodes}(\mathcal{G}, r_i)$
    \ENDFOR
    \RETURN $r_i$
\end{algorithmic}
\end{algorithm}
The \textit{Reasoner} proposes promising subgoals to search for the target object based on the \graphshort{} state representation. At a high level, it semantically reasons over the \graphshort{} $\mathcal{G}$ with LLMs to identify a subgoal, then finds a feasible path to the subgoal with graph search over $\mathcal{G}$. As with the \engineshort{}, it is a templated algorithm that is grounded into an executable routine with an \graphshort{} schema, which is described in \autoref{alg:reasoner}. It comprises 3 components (\autoref{fig:system}):

\textbf{Region Proposer.} This is a coarse-to-fine, greedy search through the layers of the \graphshort{} to propose a Place node subgoal to search. At each layer, an LLM is queried to identify the node most promising to search. This ends after iterating through all layers, with a Place node identified as subgoal, $r_{\text{subgoal}}$.

\textbf{Pathfinder.} This module extracts the connectivity subgraph of Places and Connectors from the \graphshort{}, and plans a path with Dijkstra from the robot's current location to the subgoal node returned by the Region Proposer, $r_{\text{subgoal}}$.

\textbf{Object Proposer.} This identifies objects as fine-grained subgoals for navigation. When the frontier region $r_{\text{subgoal}}$ has been reached, the Object Proposer queries LLMs to select objects in it that are likely to be close to the target. Separately, when the robot is navigating along a path to the frontier region $r_{\text{subgoal}}$, the Object Proposer is used to sample object subgoals from each region along the path. This is done because most current navigation models take in goals that are spatially localised, like objects. Thus, moving to a region is implemented as a sequence of \textsc{MoveToObject}($o$) actions, where $o$ is an Object or Connector. We follow \citet{shah2023lfg} in prompting the LLM to provide its reasoning, so as to encourage responses more relevant to the task.

\subsection{Local Navigation Controller}
\label{sec:approach_controller}

The Controller module is a low-level sensorimotor loop implementing the \textsc{MoveToObject} primitive. It maps RGB inputs to velocity commands, conditioned on a crop of a goal object. The \textsc{MoveToObject} primitive is only expected to work locally, for goal object instances that are nearby---\ie{} within field-of-view, or slightly occluded but still in the same area as the robot. Image goals are issued by the Reasoner, which grounds high-level plans to a sequence of local \textsc{MoveToObject} actions.

Recent work on image-goal visuomotor policies have shown that they form a good substrate for generalisable visual navigation across embodiments and environments~\citep{shah2023vint, sridhar2023nomad}. We build on these to implement \textsc{MoveToObject}. Concretely, we use the ViNT GNM~\citep{shah2023vint} zero-shot for visual navigation toward target objects. We approximate an object instance goal by conditioning the GNM on an image crop of the target object---taken direct from object detector, or saved in the \graphshort{}. To decide when the goal is reached, we apply a heuristic threshold based on the temporal distance metric predicted by ViNT.

\section{Experiments}
\label{sec:result}

\begin{table*}[t]
    \centering
    \begin{minipage}[b]{0.4\textwidth}
        \centering
        \caption{\textbf{ObjectNav performance of LLM ObjectNav baselines in HM3D.} We highlight the \underline{\textbf{best}} and \textbf{second-best} scores.}
        \label{tab:llm_ablations}
        \small
            \begin{tabular}{@{\extracolsep{4pt}}lccc@{}}
                \toprule
                \textbf{Method} & \textbf{SR} ($\uparrow$) & \textbf{SPL} ($\uparrow$) & \textbf{DTG}($\downarrow$) \\
                \midrule
                {FBE-GT} & 0.626 & 0.310 & 2.677 \\
                \midrule
                LGX-GT & 0.275 & 0.080 & 5.078  \\
                LFG-GT & 0.675 & \underline{\textbf{0.389}} & 2.411 \\
                \oursprgt{} & \underline{\textbf{0.775}} & \textbf{0.380} & \underline{\textbf{1.702}} \\
                \midrule
                \ourspr{} & \textbf{0.693} & 0.283 & \textbf{2.338} \\
                \bottomrule
            \end{tabular}
    \end{minipage}
    \hfill
    \begin{minipage}[b]{0.5\textwidth}
    \centering
    \footnotesize
    \caption{\textbf{Comparison with ObjectNav baselines in Gibson.} TF / NM denote training-free/non-metric approaches.}
    \label{tab:gibson_results}
        \begin{tabular}{@{\extracolsep{4pt}}lccccc@{}}
            \toprule
            \textbf{Method} & \textbf{SR} ($\uparrow$) & \textbf{SPL} ($\uparrow$) & \textbf{DTG} ($\downarrow$) & \textbf{TF} & \textbf{NM} \\
            \midrule
            SemExp & 0.657 & 0.339 & 1.474 & \xmark & \xmark  \\
            PONI & \underline{\textbf{0.736}} & \underline{\textbf{0.410}} & \underline{\textbf{1.250}} & \xmark & \xmark \\
            \midrule
            LGX & 0.310 & 0.052 & 4.775 & \cmark & \xmark\\
            LFG & 0.645 & \underline{\textbf{0.406}} & 1.812 & \cmark & \xmark\\
            FBE & 0.641 & 0.283 & 1.780 & \cmark & \xmark\\
            SemUtil & 0.693 & 0.405 & \underline{\textbf{1.488}} & \cmark & \xmark \\
            \midrule
            \ourspr{} & \underline{\textbf{0.734}} & 0.386 & 1.722 & \cmark & \cmark\\
            \bottomrule
        \end{tabular}
    \end{minipage}
\end{table*}

We evaluate \navsys{} in simulation on various ObjectNav benchmarks and also on two real robots, Fetch and Boston Dynamics Spot. We summarise our experimental findings below:

\begin{enumerate}[label=\textbf{F\arabic*}]
    \item \label{finding:gen} \textit{\navsys{} generalises zero-shot across diverse environments, embodiments and goals in the open world.}  Experiments show that foundation models enable \navsys{} to generalise across diverse environments in simulation benchmarks such as HM3D~\citep{ramakrishnan2021hm3d} and Gibson~\citep{xiazamirhe2018gibsonenv} (\autoref{sec:exp_objnav_results}) and also on real robots 
    (\autoref{sec:exp_realrobots}).  \navsys{}  also generalises over robot embodiments, both wheeled and legged robots, with different camera models and mountings (\autoref{sec:exp_gen_embodiment} and~\autoref{sec:exp_realrobots}). Finally, \navsys{} successfully handles open-vocabulary goals (\autoref{sec:exp_gen_goals}). 

    \item \label{finding:objnav} \textit{To achieve effective, generalisable object-goal navigation,  \navsys{} combines foundation models, particularly, LLMs, with \graphshort{}s}. \graphshort{}s provide a structured representation that helps LLMs to organise their rich semantic knowledge about environments for effective search (\autoref{sec:exp_objnav_results}). 
    In particular, \graphshort{} schemas support \navsys{}'s ability to generalise over different environment types. We evaluated this hypothesis on multiple LLMs, including small local models (\autoref{sec:exp_model_choice}).

    \item \label{finding:semantics} \textit{Semantics, independent of metric geometry, provides a rich source of  information for robot navigation.} \navsys{} achieves robust performance using only semantics, without explicit metric geometry (\autoref{sec:exp_objnav_results}). It uses semantic \textit{object features} for localisation, and  uses foundation models for spatial reasoning and image-goal navigation. 
    We find that object features enable surprisingly strong data association for localisation using LLMs (\autoref{sec:exp_data_assoc}).

    \item \label{finding:osg_construction} \textit{\graphshort{}s can be constructed for diverse environments in the open world.} We show that \graphshort{}s can be constructed zero-shot over diverse environments (\autoref{sec:exp_osg_building}). We further show that \graphshort{} schemas can be automatically generated for different environment types (\autoref{sec:exp_schema_autogen}). 

\end{enumerate}

\subsection{Simulation Experiments Setup}
\label{sec:sim_exp_setup}

\subsubsection{Datasets.}

We use the Habitat simulator ~\citep{szot2021habitat} with the \gibson{}~\citep{xiazamirhe2018gibsonenv} and \hmfull{}~\citep{ramakrishnan2021hm3d} datasets, both of which contain photorealistic 3D reconstructions of indoor environments. For \gibson{}, we follow \citet{ramakrishnan2022poni, chen2023semutil} to use the tiny split's validation set, comprising 1000 episodes evenly split over five household scenes (each with 12-28 rooms across 3 floors). For \hm{}, we extend the setup of \citet{shah2023lfg}, doubling the number of evaluation episodes sampled from the validation set to 400. The episodes are sampled evenly across the validation set's 20 scenes (each containing 4-19 rooms across 1-3 floors).

\subsubsection{Metrics.}

We evaluate with 3 standard metrics used in the Habitat Challenge for ObjectNav~\citep{habitatchallenge2022}:
\begin{itemize}
    \item \textit{Success rate (SR)}: The ratio of episodes where the agent succeeds, defined by reaching a position within 1.0m from the goal instance, from which the goal can be directly observed.
    \item \textit{Success-weighted path length (SPL)}. Ratio of agent's traversed distance to the ground-truth shortest path distance between the start and goal, for successful episodes. Intuitively measures efficiency of the agent's path compared to the optimal path. Given by,
    \begin{equation}
        \frac{1}{N}\sum\limits_{i=1}^{N}S_i\frac{l_i}{\text{max}(p_i, l_i)}
    \end{equation}
    $l_i, p_i$ are the optimal path length and agent's traversed distance. $S_i$ is a success indicator variable. $N$ is the total number of episodes.

    \item \textit{Distance-to-goal (DTG)}. Average distance of the agent from the goal instance at the end of each episode.
\end{itemize}

\subsubsection{Baselines.}

We compare \navsys{} to state-of-the-art ObjectNav works, and in particular analyse its benefits over recent LLM-based approaches. Baselines which do not incorporate perception and instead use ground-truth semantic annotations are marked with \textit{GT}.

We select state-of-the-art approaches which both \textbf{(i)} require ObjectNav-specific training, and \textbf{(ii)} which are training-free. For \textbf{(i)}, we evaluate with SemExp~\citep{chaplot2020semexp} and PONI~\citep{ramakrishnan2022poni}. For \textbf{(ii)}, we evaluate with purely geometric frontier-based exploration (FBE)~\citep{yamauchi1997fbe}, SemUtil~\citep{chen2023semutil} which combines FBE with semantic costs, and approaches which use LLMs to suggest or score exploration goals in the form of LGX~\citep{dorbala2024lgx} and LFG~\citep{shah2023lfg}.

LGX and LFG are ablations of \navsys{} which use less rich/structured representations compared to \graphshort{}s. In particular, we highlight the following differences among the LLM ObjectNav baselines:
\begin{itemize}
    \item \textit{LGX-GT}~\citep{dorbala2024lgx}. Prompts LLM to decide a direction to search towards, based on objects in the immediate field-of-view. \textit{I.e.} a greedy approach which does not maintain any persistent scene representation.
    \item \textit{LFG-GT}~\citep{shah2023lfg}. Prompts LLM with list of objects around the frontiers extracted from a metric map, to select promising frontiers to search. \textit{I.e.} an approach which maintains a scene representation containing information on \textit{objects} in the scene.
    \item \textit{\oursprgt{}}. Our \navsys{} system which prompts LLMs with \graphshort{}s to decide promising places to search. \graphshort{}s are richer representations that capture information on spatial regions and scene topology, in addition to objects. 
    \item \textit{\ourspr{}}. Full \navsys{} system which does not rely on ground-truth annotations, but uses VFMs for perception.
\end{itemize}

We implement the LLM baselines by adapting their publicly available source code for the ObjectNav setup. We implement a Fast Marching Method (FMM)~\citep{sethian1996fmm} controller based on \citet{chaplot2020semexp}, and standardise the baselines (LGX-GT, LFG-GT) and variants of our approach (\oursprgt{}, \ourspr{}) to use this as a local planner in simulation, to compare their performance on the basis of differences in scene representation and reasoning.

Apart from our experiments with automatically generated \graphshort{} schemas in \autoref{sec:exp_schema_autogen}, all other simulation experiments use the hand-designed schema for household environments in \autoref{listing:sim_homes}.

\subsection{Real-Robot Experiments Setup}
\label{sec:exp_rw_setup}

\begin{specbox}[specstyle, label=listing:rw_rls]{Open-plan studio apartment}
\begin{lstlisting}[language=json]
`Room`:
    ``layer_type``: "Place"
    ``layer_id``: 2
    ``has``: [Object]
    ``connects_to``: [Entrance, Room]
`Entrance`:
    ``layer_type``: "Connector"
    ``layer_id``: 2
    ``is near``: [Object]
    ``connects_to``: [Room]
`Object`:
    ``layer_id``: 1
\end{lstlisting}
\end{specbox}

We evaluate zero-shot generalisation to different robots and diverse object queries in the real world. We test in an open-plan studio apartment comprising a living room, dining room, kitchen and office, using a manually designed \graphshort{} schema described in \autoref{listing:rw_rls}. We deploy the full \navsys{} (with \engineshort{}, Reasoner, and GNM-based Controller) on two robot systems, running all modules locally on a Jetson AGX Orin, aside from queries to OpenAI's LLMs which is done over a Wi-Fi connection.

\begin{itemize}
    \item \textit{Spot}: A quadrupedal robot, with a $170^\circ$ FoV fisheye RGB camera mounted $\sim0.8$m above ground.
    \item \textit{Fetch}: A differential-drive mobile manipulator, with a $87^\circ$ FoV Realsense RGB camera mounted $\sim1.4$m above ground.
\end{itemize}

\subsection{ObjectNav in Diverse Environments}
\label{sec:exp_objnav_results}

\begin{figure*}[!t]
  \includegraphics[width=\linewidth, ]{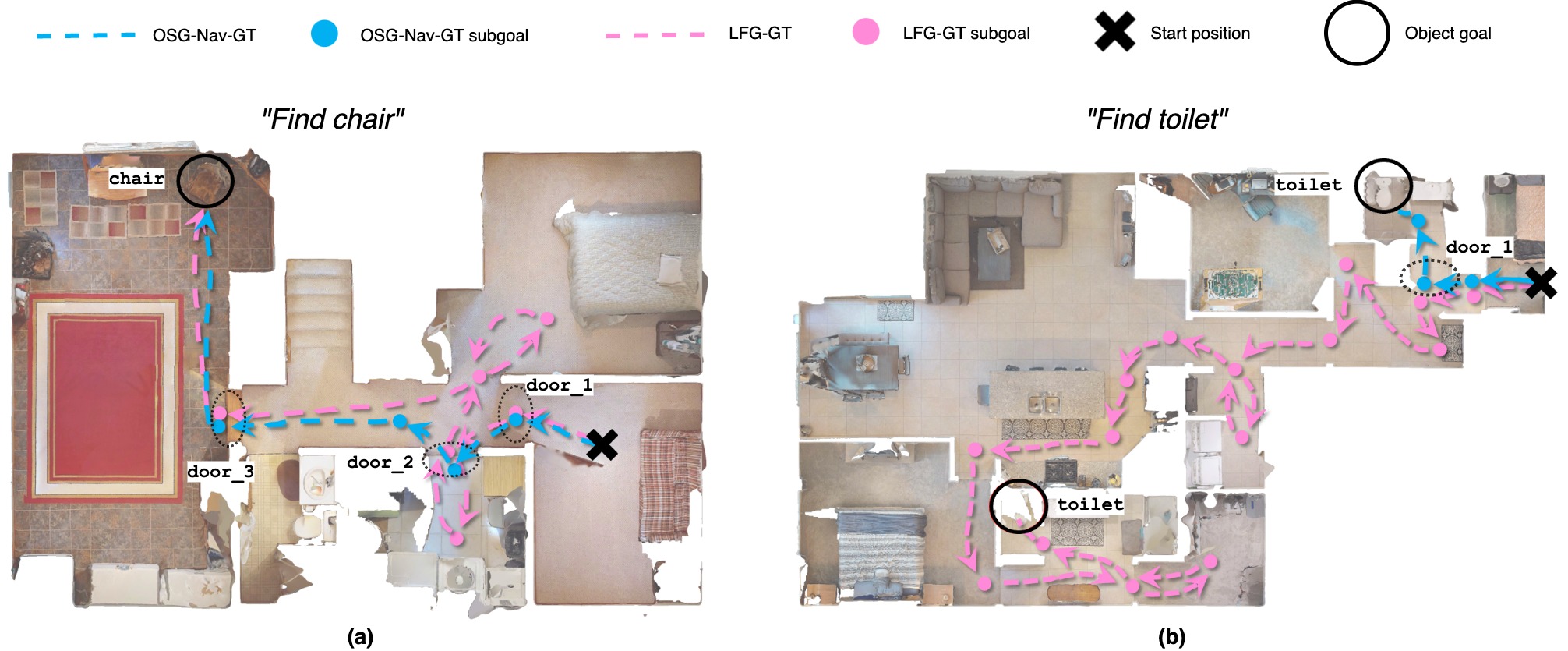}
  \caption{\textbf{Compare \oursprgt{} and LFG-GT in search efficiency.} \oursprgt{} uses region-level semantics from \graphshort{}s to search more efficiently, compared to LFG-GT which lacks a topo-semantic memory. In both scenes, the robot sees few objects at the start. \textbf{(a)} LFG-GT defaults to an exhaustive search, while \oursprgt{} quickly heads to a living room to find a chair. \textbf{(b)} \oursprgt{} sees that it starts in a bedroom and searches nearby entrances for a possible bathroom, while LFG-GT starts an exhaustive search that incidentally leads it away from the nearest toilet.}
  \label{fig:analysis}
\end{figure*}

Simulation experiments in both \hm{} (\autoref{tab:llm_ablations}) and \gibson{} (\autoref{tab:gibson_results}) support our findings that \navsys{} enables effective ObjectNav (\ref{finding:objnav}), while generalising across diverse scenes (\ref{finding:gen}). Since \navsys{} internally reasons with and uses purely semantic cues, its strong performance on a navigation task supports \ref{finding:semantics}.

Overall we find that \navsys{} has strong zero-shot performance on ObjectNav across diverse indoor environments. The \ourspr{} variant achieves a high success rate competitive with the strongest learned baselines like PONI (\autoref{tab:gibson_results}) despite not having ObjectNav-specific training. Notably, we also show stronger performance than other training-free approaches, outperforming by a wide margin SemUtil~\citep{chen2023semutil}, which combines classical planning with closed-set semantics, and other recent methods which use LLMs. The high success rates of \ourspr{} and \oursprgt{} on \hm{} point towards strong cross-environment generalisation, given the diverse collection of household scenes found in \hm{}. While our topo-semantic approach does not let us optimise for shortest geometric paths, the \navsys{} variants still remain competitive with state-of-the-art baselines in SPL and DTG. In particular, we note that we also outperform by a large margin frontier-based exploration (FBE)~\citep{yamauchi1997fbe}, a purely metric exploration approach. This highlights the value of semantic priors and reasoning for efficient search across diverse environments. 

We analyse the factors contributing to \navsys{}'s success, and find that structured representations capturing both \textit{objects} and \textit{spatial regions} enhance ObjectNav performance. From \autoref{tab:llm_ablations}, LFG-GT improves significantly over LGX-GT, highlighting the need to maintain a global scene memory for planning in ObjectNav, as opposed to greedily making choices based on current observations. \oursprgt{} outperforms LFG-GT by additionally retaining information on regions and scene topology in the \graphshort{}, suggesting the value of information on a scene's \textit{spatial structure} to reasoning for ObjectNav. 

Retaining scene information on spatial structure can enable efficient search, especially when object information is lacking. We find that \textbf{(i)} region or place semantics (\eg{} the room type) provides strong cues for reasoning, while \textbf{(ii)} \textit{Connectors} like doors and entrances are often good frontiers for exploration. In particular, an \graphshort{} schema includes \textit{Connectors} as first-class citizens to emphasise their value in reasoning for ObjectNav. \autoref{fig:analysis} highlights examples that demonstrate these benefits. In \textbf{(a)}, both LFG-GT and \oursprgt{} face a scene with multiple rooms branching off a main corridor, and with few objects in sight. The emphasis on Connectors focuses \oursprgt{} on searching through new doors in the scene, while its ability to explicitly identify rooms and understand their semantic utility enables it to quickly eliminate new regions that are unsuitable for exploration. In contrast, LFG-GT defaults to an inefficient exhaustive search over the scene. Similarly in \textbf{(b)}, both LFG-GT and \oursprgt{} begin in a sparsely furnished bedroom. \oursprgt{} is able to identify its surroundings and reasons that toilets are commonly found in the vicinity of bedrooms, leading it to focus its search on nearby doors to efficiently locate the toilet. In contrast, LFG-GT defaults to frontier exploration on a metric map, selecting suboptimal object frontiers that lead it away from the nearest toilet.

\subsection{Generalisation over Open-Vocabulary Goals}
\label{sec:exp_gen_goals}

\begin{table*}
    \begin{minipage}[b]{0.37\textwidth}
    \centering
    \tiny
    \caption{\textbf{Real-world, open-world ObjectNav.}}
    \label{tab:real_world_exp}
    \small
        \begin{tabular}{@{\extracolsep{4pt}}lcc@{}}
            \toprule
            \textbf{Object goal} & \textit{Spot} & \textit{Fetch} \\
            \midrule
            \texttt{guitar} & 4/5 &  4/5 \\
            \texttt{dish washer} & 5/5 &  4/5 \\
            \bottomrule
        \end{tabular}
    \end{minipage}
    \begin{minipage}[b]{0.63\textwidth}
    \centering
    \scriptsize
    \caption{\textbf{Performance of \ourspr{} and LGX on rare object queries.}}
    \label{tab:rare_goals}
        \begin{tabular}{@{\extracolsep{4pt}}lcccccc@{}}
        \toprule
         & \multicolumn{3}{c}{\textbf{Settee}} & \multicolumn{3}{c}{\textbf{Nightstand}} \\ \cmidrule{2-4} \cmidrule{5-7}
          \textbf{Method}& \textbf{SR} ($\uparrow$) & \textbf{SPL} ($\uparrow$) & \textbf{DTG}($\downarrow$)     & \textbf{SR} ($\uparrow$) & \textbf{SPL} ($\uparrow$) & \textbf{DTG}($\downarrow$)     \\  \midrule
        LGX~\citep{dorbala2024lgx} & 0.1   & 0.023  & 6.186  & 0.4     & 0.081     & 2.873    \\ \midrule
        \ourspr{}     & 0.8    & 0.285 & 0.894   & 0.7     & 0.287     & 1.699\\
        \bottomrule
        \end{tabular}
        \end{minipage}
\end{table*}

\begin{table*}[tbp]
\caption{\textbf{Success rate of \ourspr{} in handling open-vocabulary compositional queries.} Queries are object goals composed with one or more phrases describing colour (C), material (M) or pattern (P). \textit{E.g.}, a ``red, floral bed'' (C+P with goal ``bed'').}
\footnotesize
    \begin{center}
        \begin{tabular}{@{\extracolsep{4pt}}lcccccccccc@{}}
            \toprule
            \textbf{}&\textbf{}&\textbf{}&\textbf{}&\multicolumn{7}{c}{Success rate (over 10 trials each)}\\
            \cmidrule{5-11}
            \textbf{Object goal} & \textbf{Colour (C)} & \textbf{Material (M)} & \textbf{Pattern (P)} & \textbf{C} & \textbf{M} & \textbf{P} & \textbf{C+M} & \textbf{C+P} & \textbf{M+P} & \textbf{C+M+P}\\
            \midrule
            \texttt{chair} & \textit{black and white} & \textit{fabric} & \textit{leopard-print} & 0.8 & 0.6 & 0.8 & 0.6 & 0.6 & 0.6 & 0.6 \\
            \texttt{bed} & \textit{red} & \textit{cotton} & \textit{floral} & 0.8 & 0.8 & 0.8 & 0.6 & 0.6 & 0.6 & 0.4\\
            \bottomrule
        \end{tabular}
    \label{tab:compositional_goals}
    \end{center}
\end{table*}

We find that \navsys{} has a strong ability to handle diverse open-vocabulary goals (\ref{finding:gen}). Our real-world experiments (\autoref{tab:real_world_exp}) show that \navsys{} is capable of locating open-set goals like ``guitar'' and ``dish washer'', which are not specified goal classes in ObjectNav~\citep{habitatchallenge2022}. We provide a more detailed analysis of \navsys{}'s open-vocabulary abilities by evaluating on \textit{rare queries} and \textit{compositional queries} in simulation. The former are uncommon, long-tailed object concepts, which we select based on the LVIS dataset~\citep{gupta2019lvis}. The latter are complex language descriptions of an object at different levels of granularity, which compose an object noun (\eg{} ``chair'') with descriptors (\eg{} ``leopard-print'', ``fabric'' \etc{}). While the individual nouns and adjectives may be common, the resulting queries are overall uncommon and unlikely to be present in the models' training data. We design new navigation episodes in \hm{} for these queries.

We find that \navsys{} generalises well over \textit{rare object queries}. To evaluate these we select two uncommon nouns based on LVIS: \texttt{settee}, an uncommon synonym for the commonplace couch; \texttt{nightstand}, an object uncommon enough that it is excluded from LVIS. \autoref{tab:rare_goals} shows strong performance when queried with these rare nouns, suggesting strong generalisation to rare object queries. To evaluate \textit{compositional object queries}, we compose selected object nouns with one or more phrases to describe object properties like colour, material or pattern. The navigation episodes designed for these require \navsys{} to disambiguate amongst different object instances based on the provided descriptors (\eg{} differentiating a ``leopard-print'' chair from a ``floral-print'' chair). From \autoref{tab:compositional_goals}, we find that \navsys{} performs well in general on compositional queries. We note slight drops in success rate as descriptors are added and the requirements on object appearance become more complex. We attribute this to perception errors which misclassify the object properties and cause \navsys{} to fail to recognise the goal.

\begin{figure}[!t]
    \centering
    \includegraphics[width=\linewidth]{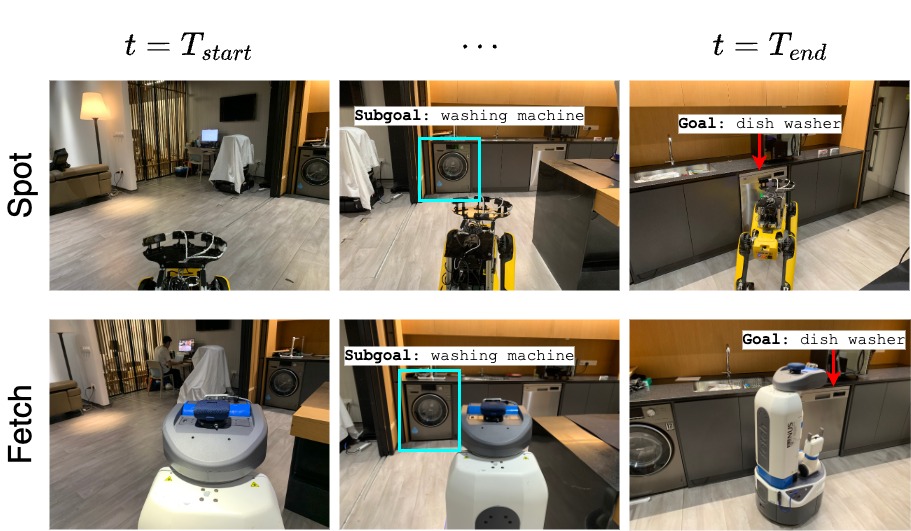}
    \caption{\textbf{Cross-embodiment generalisation in the real world.} \navsys{} is deployed zero-shot across two different robots, Spot and Fetch, without any parameter tuning. It perceives, reasons and acts similarly despite the change in embodiment and sensors, successfully reaching the goal dish washer in both cases.}
    \label{fig:real_world_exp}
\end{figure}

\subsection{Generalisation over Robot Embodiments}
\label{sec:exp_gen_embodiment}

We evaluate cross-embodiment generalisation through real world experiments on different robots (\autoref{fig:real_world_exp}), supporting our findings in \ref{finding:gen}. These robots feature different dynamics (quadrupedal vs. differential drive) and different sensor inputs (different mounting heights, fisheye vs pinhole cameras). \autoref{tab:real_world_exp} shows that \navsys{} succeeds often on various ObjectNav tasks across both robots in spite of the widely varying inputs and dynamics. This is enabled by composing \navsys{} fully from foundation models which have shown broad generalisation across perceptual inputs (VFMs) and robot dynamics (GNMs). We note that the GNM tends to perform slightly worse at goal-reaching when narrow field-of-view cameras like the Realsense are used, and that its performance on goal-directed navigation in cluttered areas is limited: these issues induced the observed failures in the real world experiments.

\subsection{Object Features for Data Association}
\label{sec:exp_data_assoc}

\begin{table*}[t]
    \centering
    \begin{minipage}[b]{0.49\textwidth}
    \centering
    \caption{\textbf{Accuracy of data association with object features.} We test LLMs' ability to use object features to \textit{recognise} the same scene element from different views, or \textit{distinguish} different elements (\eg{} two different Objects, or two different Places).}
    \label{tab:obj_feat}
    \small
        \begin{tabular}{@{\extracolsep{4pt}}lccc@{}}
            \toprule
            \textbf{Node} & \textbf{Recognise} & \multicolumn{2}{c}{\textbf{Distinguish}} \\
            \cmidrule{3-4}
            & & \textit{Same type} & \textit{Diff. type} \\
            \midrule
            Objects & 0.85 & 0.58 & - \\
            Connectors & 0.75 & 0.83 & - \\
            Places & 0.88 & 0.80 & 0.94 \\
            \bottomrule
        \end{tabular}
    \end{minipage}
    \hfill
    \begin{minipage}[b]{0.49\textwidth}
    \centering
    \caption{\textbf{Evaluation of scene graph quality.} We measure \textbf{Pr}ecision and \textbf{Re}call in the estimation of nodes and outgoing edges for \textit{floors} (Region Abstractions), \textit{rooms} (Places), \textit{doors} (Connectors)}
    \label{tab:scene_graph_quality}
    \scriptsize
        \begin{tabular}{@{\extracolsep{4pt}}lcccccc@{}}
            \toprule
            \textbf{}&\multicolumn{2}{c}{Nodes}&\multicolumn{2}{c}{\edgecontains{}}&\multicolumn{2}{c}{\edgeconnects{}} \\
            \cmidrule{2-3}\cmidrule{4-5}\cmidrule{6-7}
            \textbf{} & \textbf{Pr} & \textbf{Re} & \textbf{Pr} & \textbf{Re} & \textbf{Pr} & \textbf{Re} \\
            \midrule
            Floors  & 1.000 & 1.000 & 0.889 & 0.914 & 1.000 & 1.000   \\
            Rooms & 0.846 & 0.880 & - & - & 0.771 & 0.831 \\
            Doors &0.845 & 0.800 & - & - & 0.783 & 0.857 \\
            \bottomrule
        \end{tabular}
    \end{minipage}
\end{table*}

As highlighted in \ref{finding:semantics}, semantics alone are a rich source of information, enabling key functions in navigation that traditionally rely on geometric information. Specifically, we find that purely semantic \textit{object features} are surprisingly effective for associating regions and objects. Such data association is essential for place recognition and object re-identification, and usually requires dense visual descriptors or geometric features.

This section evaluates the accuracy of object features in associating Objects, Connectors and Places. To do so, we use a dataset of views sampled from \hm{} that is curated to examine how well object features enable LLMs to \textit{recognise} the same Object/Connector/Place from different views and to \textit{distinguish} among different Objects/Connectors/Places. The dataset is constructed by sampling rooms across 10 randomly selected \hm{} scenes, then sampling views within selected rooms. We ensure that there are multiple views of each labelled object, and also ensure that multiple rooms of the same function and type (\eg{} multiple bedrooms) are sampled, to make the task realistic and challenging.

From \autoref{tab:obj_feat}, we find that object features reliably enable LLMs to recognise and distinguish Objects, Connectors and Places in the environment. We note that the tests to distinguish Objects and Connectors involve two distinct Object/Connector instances with the same type or semantic label (\eg{} two instances of ``chairs''). We find that LLMs using object features can struggle to correctly distinguish distinct object instances of the same type, as human environments often have similar objects clustered closely (\eg{} multiple chair instances grouped closely around a table). In these cases, each object's neighbours (and hence features) are too similar, causing aliasing. Practically, such incorrect associations do not significantly affect ObjectNav, since erroneously identified objects are close to the correct object, and the system will navigate to the vicinity of the correct object. Additionally, \autoref{tab:obj_feat} also shows that LLMs can strongly distinguish Places of different types (\eg{} bedroom and kitchen) purely using the objects contained within them. This suggests that object features provide a robust source of localisation information to augment place classification-based localisation.

\begin{figure}[!t]
    \centering
    \includegraphics[width=\linewidth]{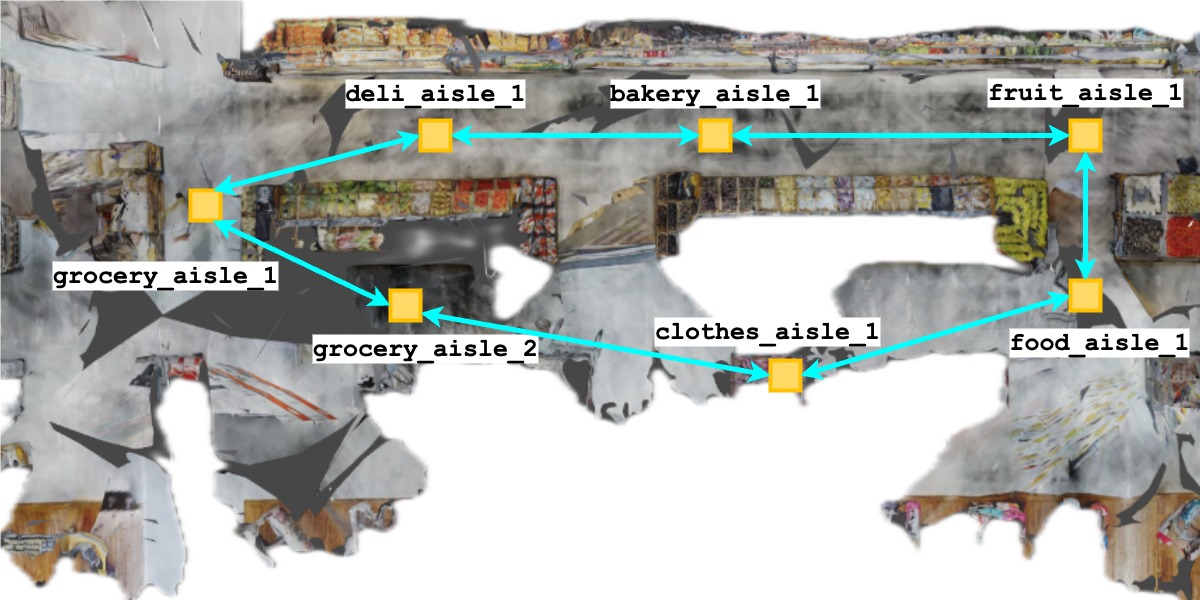}
    \caption{\textbf{Generated \graphshort{} for supermarket.} The \engineshort{} constructs an \graphshort{} of the Gratz supermarket scene from the Gibson dataset using \autoref{listing:osg_spec_supermarket} (in \autoref{app:exp_mapper_rw}), which uses ``aisles'' as Place nodes.}
    \label{fig:osg_supermarket}
\end{figure}

\subsection{Online \graphshort{} Construction}
\label{sec:exp_osg_building}

\begin{figure*}[!t]
    \centering
    \includegraphics[width=0.85\linewidth]{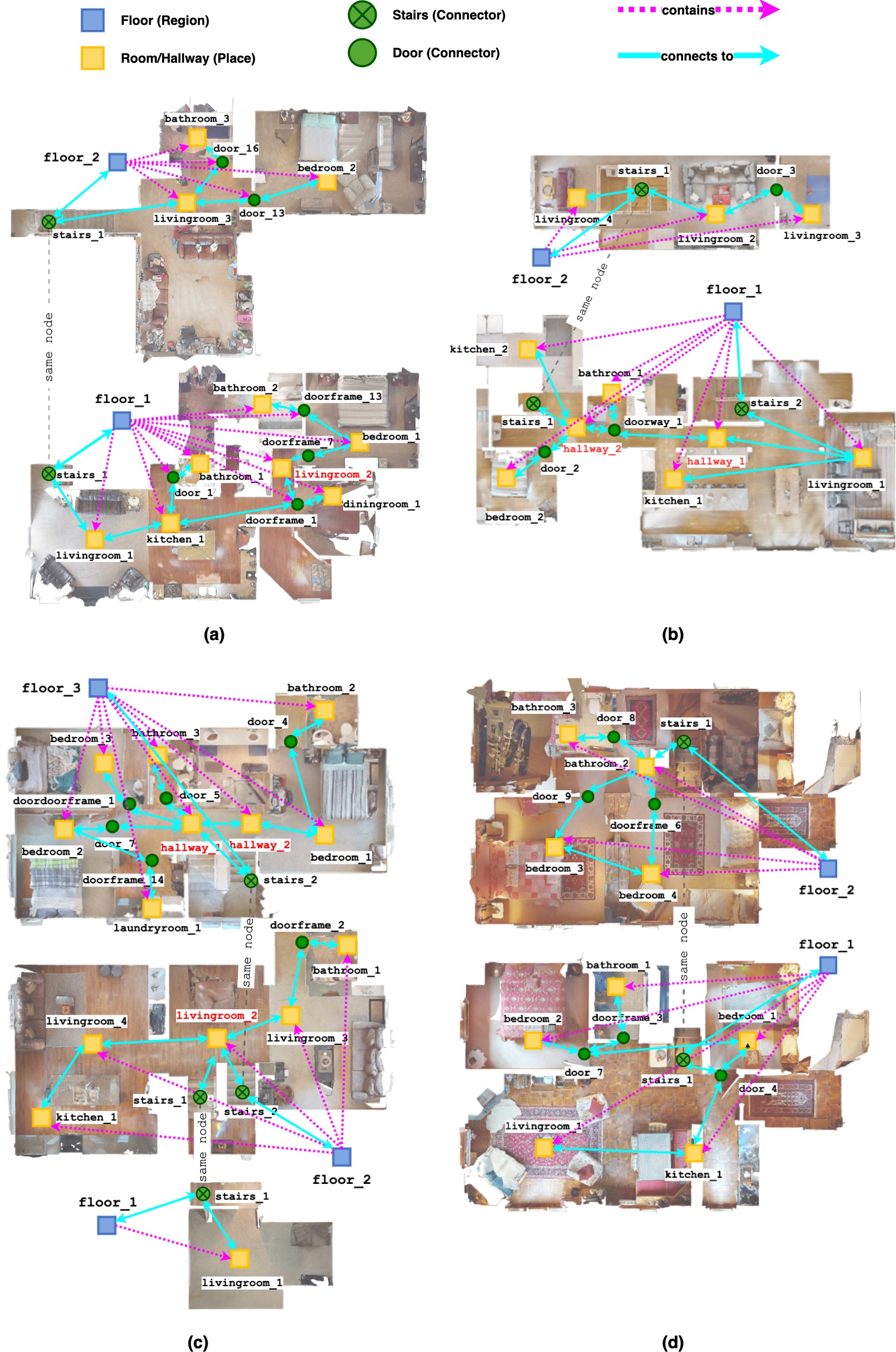}
    \caption{\textbf{Generated \graphshort{}s for multi-storey homes.} We use \engineshort{} to construct \graphshort{}s using sensor data from teleoperated trajectories from \hm{} scenes. The \graphshort{}s are built with \autoref{listing:sim_homes}. The resulting \graphshort{}s accurately capture Places (rooms/hallways), Connectors (doors/stairs), and Regions (floors), along with their spatial connectivity and hierarchy. The main failure modes observed are oversegmenting of hallways in \textbf{(b), (c)} and misclassifications by the BLIP2 VQA model---identifying a hallway as living room in \textbf{(a)}, and a dining room as a living room in \textbf{(c)}. The errors are highlighted in \textcolor{red}{red}.}
    \label{fig:osg_multifloor}
\end{figure*}

\begin{figure}[!t]
    \centering
    \includegraphics[width=\linewidth]{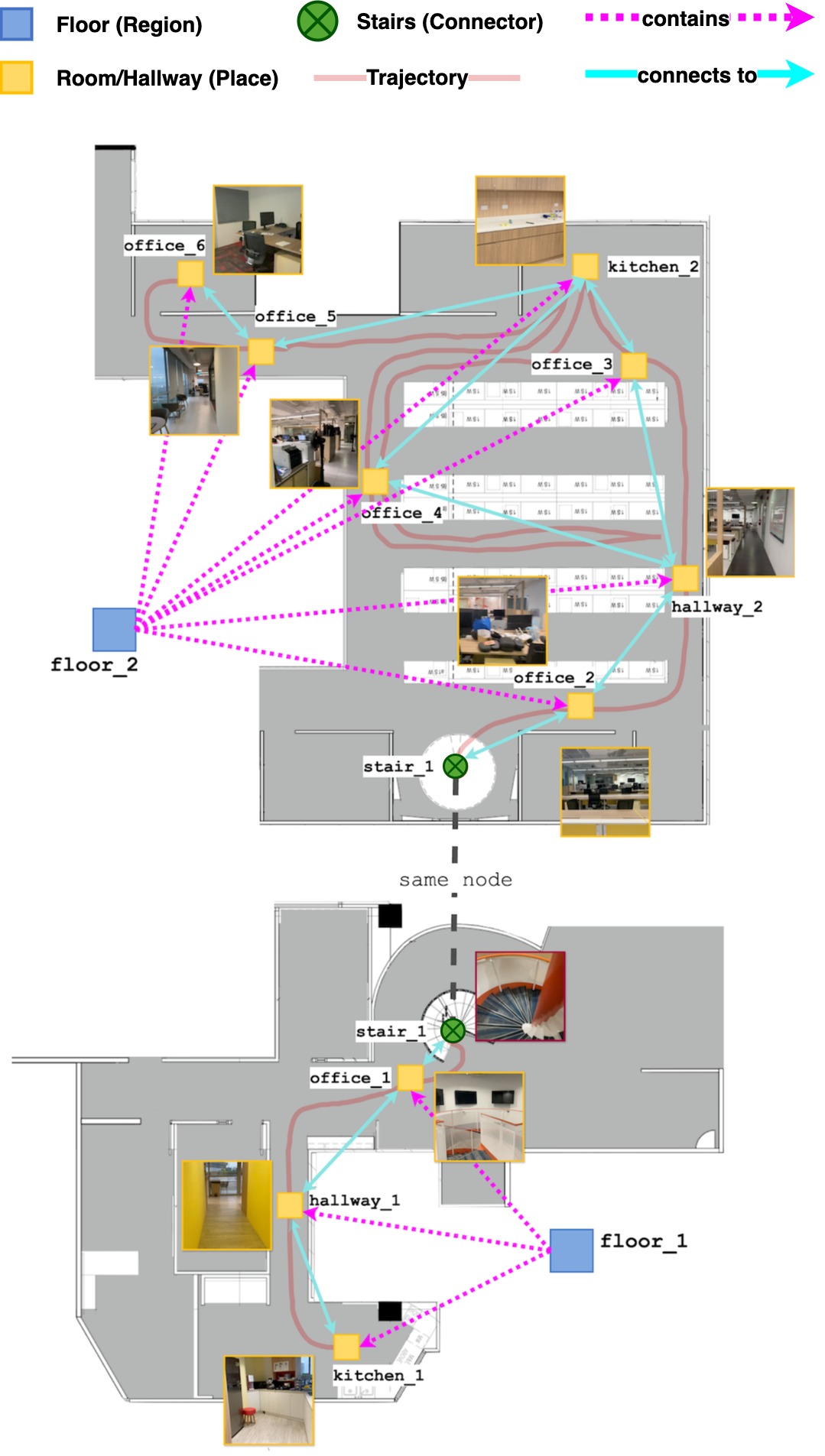}
    \caption{\textbf{Generated \graphshort{} for a real-world office scene.} The \graphshort{} is constructed using data from the \textcolor{red}{red} trajectory, and includes the Place concepts of \textit{office}, \textit{room}, \textit{hallway}, and the Connector \textit{stairs} (\autoref{listing:osg_spec_office}, in \autoref{app:exp_mapper_rw}).}
    \label{fig:osg_real_world_office}
\end{figure}

We show that the \engineshort{} can accurately map and capture the spatial structure of scenes across varied environment types, supporting \ref{finding:osg_construction}. First, we evaluate the accuracy of \graphshort{} construction over five different homes in \hm{} by comparing the generated \graphshort{}s to human-annotated scene graphs. Second, we provide qualitative examples of \graphshort{}s built across different environment types in both simulation and the real world.

\autoref{tab:scene_graph_quality} shows \graphshort{} quality in terms of the precision and recall of \engineshort{} in predicting \graphshort{} nodes and their outgoing edges. We use human-annotated graphs as ground truth, and consider Connector (door), Place (room) and Region Abstraction (floor) nodes, along with the edges connecting them, in our evaluation. We observe strong performance in general across all tests and metrics, indicating that most \graphshort{}s generated by the \engineshort{} accurately capture the scene's spatial structure and topology. We show examples of \graphshort{}s generated from these experiments in \autoref{fig:osg_multifloor}. We note that the \graphshort{} successfully identifies and labels most of the spatial regions (i.e. \textit{rooms}, \textit{hallways}, \textit{floors}) in the scene, and also accurately captures the connectivity and hierarchy of these regions. The mapper can sometimes oversegment long hallways into several smaller hallway regions, which does not significantly affect the topology of the \graphshort{}. Another key failure mode comes from occasional mislabelling of the Place nodes by the BLIP2 VQA model, an issue that can likely be addressed through the use of more powerful VLMs, \eg{} LLaVA \citep{liu2023visualinstructiontuning}.

We demonstrate the zero-shot applicability of the \graphshort{} framework to novel environment types, like a supermarket from the \gibson{} dataset (\autoref{fig:osg_supermarket}). We make no modifications to the \engineshort{}, only providing an \graphshort{} schema for supermarkets that specifies ``aisles'' as a suitable Place abstraction. The resulting \graphshort{} is able to decompose the scene into aisles, and accurately identifies the \textit{type} or semantic utility of each aisle (\eg{} marking an aisle for deli items, or for clothing). 

We extend our tests to the real world, building \graphshort{}s for novel environments like an \textit{office} or \textit{mall}. This requires the \engineshort{} to handle diverse spatial abstractions, ranging from offices/hallways in office environments to stores/walkways in malls. We find that it is largely able to handle these abstractions well and generate \graphshort{}s which accurately represent these scenes. For instance, \autoref{fig:osg_real_world_office} shows an \graphshort{} of a real-world office scene, which largely captures the connectivity of the environment based on sensor data from the traversed trajectory. In particular, \engineshort{} succeeds at a performing an object feature-based loop closure at \texttt{hallway\_2}. \autoref{app:exp_mapper_rw} provides the schema used for this scene, along with further examples.

\subsection{\graphshort{} Schema Generation}
\label{sec:exp_schema_autogen}

\begin{table}
    \centering
    \caption{\textbf{Performance of \ourspr{} with automatically generated \graphshort{} schemas for homes.} Results on 10 \hm{} ObjectNav episodes with \navsys{}, using    schemas generated with different methods: \textbf{(i)} manually specified by user, \textbf{(ii)} automatically generated by directly prompting LLM, \textbf{(iii)} automatically generated with our pipeline (\autoref{sec:approach_auto_spec_gen}).}    
    \label{tab:auto_spec}
    \scriptsize
        \begin{tabular}{@{\extracolsep{5pt}}lcccc@{}}
            \toprule
            \textbf{\graphshort{} spec} & \textbf{Valid schema SR} ($\uparrow$) & \textbf{SR} ($\uparrow$) & \textbf{SPL} ($\uparrow$) & \textbf{DTG} ($\downarrow$) \\
            \midrule
            Manual & - & 0.7  & 0.241 & 1.110 \\
            \midrule
            Auto: LLM & 0.0 & 0.0 & -  & -\\
            Auto: Ours & 1.0 & 0.6 & 0.220  & 2.759 \\
            \bottomrule
        \end{tabular}
\end{table}

We demonstrate our pipeline's ability to generate meaningful \graphshort{} schemas, that are also useful for ObjectNav (\ref{finding:osg_construction}). We do so by qualitatively assessing schemas generated across various human environments of differing complexity---homes, hospitals and airports--- and validating the performance of generated home schemas on ObjectNav in selected \hm{} scenes.

Our system produces schemas that are \textit{correct}, but which tend to have limited \textit{coverage} in complicated environments. The schemas are correct in that the generated spatial concepts are relevant to the prompted environment (\eg{} \textit{rooms} in homes, \textit{wards} in hospitals, \textit{gates}/\textit{terminals} in airports), and these concepts are correctly organised hierarchically (\eg{} \textit{gates} contained within \textit{terminals}). Our system also ensures good coverage in simple environments like apartment homes, \ie{} the schemas are sufficiently broad and descriptive enough to apply to such environments. We attribute this to LLMs' strong ability to capture mainstream spatial concepts from Internet-scale text. However, we find that our system may omit more specialised region concepts---\eg{} including \textit{wards} but leaving out \textit{operating theatres} in hospitals. It also tends to omit regions not directly related to the environment's function---\eg{} leaving out \textit{shops} from airports. We find that these issues can be addressed by adding more user-provided guidance: \eg{} extending the prompt from ``airport'' to ``airport containing shopping areas''. Future work could also explore the design of an interactive language-based interface by which humans can incorporate richer priors and guidance into the schema generator.

We validate that the generated schemas enable effective \graphshort{} construction and ObjectNav on  selected \hm{} episodes. Specifically, we automatically generate schemas for apartment homes at the start of each ObjectNav episode, using our Generate-Canonicalise-Verify pipeline. \autoref{tab:auto_spec} shows that our generated schemas enables performance comparable to manually designed ones. The main observed failure mode was that the generated schema sometimes left out optional Connector (\textit{door}) nodes. While suitable for open-plan apartments, this prevented it from accurately capturing spatial connectivity in household scenes with multiple doors. We also find our iterative approach to be essential for generating valid schemas. In contrast, directly prompting LLMs to do so yielded malformed JSON schemas that often did not satisfy \graphshort{} structural requirements.

\begin{specbox}[specstyle, label=listing:auto_home]{Apartment home (Auto-generated)}
\begin{lstlisting}[language=json]
`Home`:
     ``layer_type``: "Region"
     ``layer_id``: 3
     ``contains``: [Room]
`Entrance`:
     ``layer_type``: "Connector" 
     ``layer_id``: 2
     ``is near``: [Object]
     ``connects_to``: [Room]
`Room`:
     ``layer_type``: "Place"
     ``layer_id``: 2
     ``has``: [Object]
     ``connects_to``: [Entrance]
`Object`:
    ``layer_id``: 1
\end{lstlisting}
\end{specbox}

\begin{specbox}[specstyle, label=listing:auto_hospital]{Hospital (Auto-generated)}
\begin{lstlisting}[language=json]
`Hospital`:
     ``layer_type``: "Region"
     ``layer_id``: 3
     ``contains``: [Ward]
`Corridor`:
     ``layer_type``: "Connector" 
     ``layer_id``: 2
     ``is_near``: [Object]
     ``connects_to``: [Ward]
`Ward`:
     ``layer_type``: "Place" 
     ``layer_id``: 2
     ``has``: [Object]
     ``connects_to``: [Corridor]
`Object`:
    ``layer_id``: 1
\end{lstlisting}
\end{specbox}

\begin{specbox}[specstyle, label=listing:auto_airport]{Airport (Auto-generated)}
\begin{lstlisting}[language=json]
`Terminal`:
     ``layer_type``: "Region"
     ``layer_id``: 3
     ``contains``: [Gate]
     ``connects_to``: [Terminal]
`Gate`:
     ``layer_type``: "Place"
     ``layer_id``: 2
     ``has``: [Object]
     ``connects_to``: [Gate]
`Object`:
    ``layer_id``: 1

\end{lstlisting}
\end{specbox}

\subsection{Choice of Foundation Models}
\label{sec:exp_model_choice}

We analyse the role of LLM choice on the effectiveness of ObjectNav in \autoref{tab:fm_ablations}, to provide deeper understanding for \ref{finding:objnav}. LLMs are essential to \navsys{} for semantic reasoning and belief updating, but can be costly to run. We therefore explore the performance of \navsys{} on smaller models that enable efficient local inference, \ie{} the LLaMA 2~\citep{touvron2023llama} and 3~\citep{dubey2024llama3} family of models. For all experiments with local models, we run the LLMs on a single Nvidia A5000 GPU. We find that performance is halved when using the 7B LLaMA 2 variant, owing to \textbf{(1)} its poor performance in predicting stop tokens, causing infinite recursion; \textbf{(2)} its tendency to reply to the prompts used in our state estimator with ``cannot decide''; \textbf{(3)} mistakes in classifying detections into Objects, Connectors and Places when updating the \graphshort{}. These challenges are mitigated with the 13B variant. While performance remains markedly poorer than GPT-3.5, we believe these smaller models offer a reasonable tradeoff, potentially enabling \navsys{} to run fully locally and improve efficiency.

\begin{figure}[!t]
  \includegraphics[width=\linewidth]{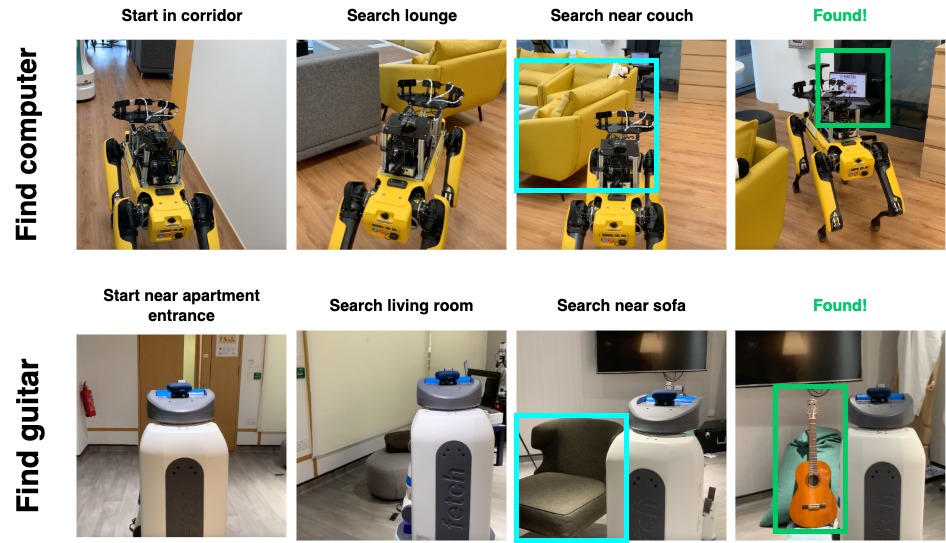}
  \caption{\textbf{Real-world ObjectNav.} The Spot robot searches common areas in a nursing home to find a computer (upper row), while the Fetch robot searches an apartment to find a guitar (lower row).}
  % \caption{\textbf{Real world case studies.} Upper row: the Spot robot is tasked to find a computer in a mock-up nursing home environment, and navigates through the corridor into the lounge, where it finds the computer near a table behind a couch. Bottom row: the Fetch robot is tasked to find a guitar, and starts near the entrance and navigates into the living room, where it locates the guitar.}
  \label{fig:exp_rw_case_studies}
\end{figure}

\begin{table}
    \centering
    \caption{\textbf{Experiments on \ourspr{}'s ObjectNav performance using different LLMs.} Results are based on 100 episodes, with 5 episodes sampled from each scene in the \hm{} validation set.}
    \label{tab:fm_ablations}
    \scriptsize
        \begin{tabular}{@{\extracolsep{4pt}}cccc@{}}
            \toprule~
            \textbf{LLM} & \textbf{SR} ($\uparrow$) & \textbf{SPL} ($\uparrow$) & \textbf{DTG} ($\downarrow$) \\
            \midrule
            Meta LLaMA 2 (7B)~\citep{touvron2023llama} & 0.349 & 0.095 & 4.190  \\
            Meta LLaMA 3.1 (8B)~\citep{dubey2024llama3} & 0.492 & 0.145 & 3.241\\
            Meta LLaMA 2 (13B)~\citep{touvron2023llama} & 0.538 & 0.124 & 2.867\\
            OpenAI GPT-3.5 &\textbf{0.693} & \textbf{0.282} & \textbf{2.385} \\
            \bottomrule
        \end{tabular}
\end{table}

\subsection{Real-Robot Experiments}
\label{sec:exp_realrobots}

We support \ref{finding:gen} with additional real-world experiments in an open-plan studio and nursing home mock-up (\autoref{fig:exp_rw_case_studies}). As with our previously described real-world tests, the \navsys{} is deployed zero-shot on different robots (\autoref{sec:exp_rw_setup}) without any change in model weights or parameters from the simulation tests. We find that it is generally robust to shifts in environment and visual appearance, and demonstrates the ability to perform ObjectNav and locate open-vocabulary objects outside the usual ObjectNav categories, like a guitar and a computer. The supplementary video contains these and other tests, along with a demonstration of topo-semantic SLAM using \graphshort{}s in the real world.

VFMs' strong generalisation over environments and across camera models and mountings are a key enabler of this robust transfer. While VFMs overall perform well in zero-shot settings, we find that they still remain sensitive to specific environmental conditions (\eg{} strong illumination changes). We also note that open-set object detectors like GroundingDINO appear to trade off some detection accuracy compared to specialised detectors trained on closed-set object detection with a much more limited vocabulary, a phenomenon also noted by \citet{chang2023goat}.

Generally, we find that GNMs are the main factor limiting the \navsys{}'s ObjectNav performance in the real world. Firstly, while GNMs are capable of reactive obstacle avoidance, their capacity for goal-directed visual navigation is limited. We attribute this to limited local planning ability, causing GNMs to get stuck in local minima in slightly more cluttered scenes. Our experiments take place in more open, cleaner environments, reducing the occurrence of such issues. We view this as an important issue for practical implementation, which can be addressed through adding heuristic recovery behaviours, sampling exploration trajectories from a GNM to escape local minima as done in ~\citet{sridhar2023nomad}, and through future work on GNMs. Secondly, GNMs ability to handle different sensing modalities is notably more limited than those of VFMs. We find empirically that GNMs have a clear bias toward wider field-of-view cameras, making more dynamic movements and taking tighter turns with them. The authors attribute this to the composition of the dataset for training GNMs.

%===============================================================================

\section{Conclusion and Future Work}
\label{sec:conclusion}

This paper presents \navsys{}, a neuro-symbolic system that combines foundation models and the \graphshort{} representation in a modular architecture for open-world  {ObjectNav}. \graphshort{} is a novel scene graph representation that leverages  \textit{schemas} to enable effective generalisation over diverse  environments. Experiments demonstrate that \navsys{} achieves state-of-the-art performance on ObjectNav benchmarks, while generalising across environments, robots and open-vocabulary queries. They further indicate that \graphshort{}s provide the rich, structured representation of the scene necessary to enable such performance.

%% Limitations
Our approach currently relies only on topo-semantic information. However, incorporating metric information---\eg{} positioning information like wheel odometry, or geometric information like depth from foundation models~\citep{yang2024depthanything}---can enhance understanding of rich, 3D scenes. This extends to providing finer-grained geometric understanding needed for higher precision tasks like manipulation.

While our work leverages \graphshort{} schemas for cross-environment generalisation, it does not yet fully exploit their flexibility. These schemas can be adapted online to better suit the environments being navigated, improving performance across episodes in complex or heterogeneous spaces (\eg{}, a transport hub linked to a shopping centre). This is reminiscent of ``iterative'' vision-and-language navigation~\citep{krantz2023iterative} which emphasises lifelong learning, and developing a similar benchmark for our setting is a promising direction. Although we focus on indoor spaces, the generic ontology can describe outdoor, human-built environments as well. However handling large-scale, open-plan areas may require advances in our mapping pipeline. Outdoor spaces are often more heterogeneous than indoor spaces, potentially requiring more complex schemas to describe. Lastly, our current ontology targets spatial regions for navigation but could be extended to support finer-grained regions relevant to manipulation tasks.

The \navsys{} currently uses LLMs in a limited capacity. It relies on very large, cloud-based LLMs like GPT-3.5, and does not fully exploit their multilingual and multimodal capabilities. We highlight that ongoing research into building compact, efficient LLMs~\citep{deitke2024molmo}, along with promising preliminary results with smaller LLaMA models (\autoref{sec:exp_model_choice}) offer positive indications that \navsys{} can be optimised for more constrained, fully onboard compute. We also leave to future work the harnessing of LLMs' multilingual and multimodal capabilities, which can enable richer interaction with users.

\begin{acks}
% This class file was developed by Sunrise Setting Ltd,
% Brixham, Devon, UK.\\
% Website: \url{http://www.sunrise-setting.co.uk}

This research is supported by the DSO National Laboratories' Graduate Fellowship.
\end{acks}

\bibliographystyle{SageH}
\bibliography{references}

\clearpage
\appendix
\section{\enginelong{} details}
\label{app:mapper_details}

\subsection{Image Parser prompts}

\begin{vqapromptbox}[specstyle, label=listing:place_class_struct]{Prompt for Identifying Place Type}
\begin{lstlisting}[style=Prompt]
Which of the place types [|<PlaceClass1>|, |<PlaceClass2>|, ... ] best describes the scene?
\end{lstlisting}
\end{vqapromptbox}

\begin{vqapromptbox}[specstyle, label=listing:place_label_struct]{Prompt for Place Labelling}
\begin{lstlisting}[style=Prompt]
What |<PlaceClass>| are we in?
\end{lstlisting}
\end{vqapromptbox}

\begin{vqapromptbox}[specstyle, label=listing:obj_description_struct]{Prompt for Object/Connector Appearance Description}
\begin{lstlisting}[style=Prompt]
What color is the |<object>|/|<connector>| in the image?

What material is the |<object>|/|<connector>| made of?
\end{lstlisting}
\end{vqapromptbox}

\begin{llmpromptbox}[specstyle, label=listing:classify_elements_struct]{Templated prompt for Scene Element Classification}
\begin{lstlisting}[style=Prompt]
@@Few-shot prompts@@
(*@\textbf{<Few-shot prompts follow format of query shown below. Examples of few-shot prompts given in \autoref{listing:classify_elements_struct}>}@*)

@@Query@@
We observe the following: ["livingroom_0", "window_13", ...]. Please eliminate redundant strings in the element from the list and classify them into |<PlaceClass>|, |<ConnectorClass>|, Object classes.
**Answer**: 
|<PlaceClass>|: ... 
|<ConnectorClass>|: ...
**object**: ...
\end{lstlisting}
\end{llmpromptbox}

\begin{llmpromptbox}[specstyle, label=listing:classify_elements_ex]{Example interaction with LLM for Scene Element Classification}
\begin{lstlisting}[style=Prompt]
@@Few-shot prompts@@
There is a list: ["bathroom_0", "bathroom mirror_1","bathroom sink_2","toilet_3", "bathroom bathtub_4", "lamp_5", "ceiling_10"]. Please eliminate redundant strings in the element from the list and classify them into "room," "entrance," and "object" classes. Ignore floor, ceiling and wall.

**Answer**: 
room: bathroom_0
entrance: none
object: mirror_1, sink_2, toilet_3, bathtub_4, lamp_5

...

@@Example query@@
We observe the following: ["livingroom_0", "window_13","door_2", "doorway_3", "table_4","chair_5","livingroom sofa_6", "floor_7", "wall_8", "doorway_9", "stairs_10", "tv_16", "stool_17", "couch_18", "remote_19"]. Please eliminate redundant strings in the element from the list and classify them into room, doorway, object classes.

(@Example LLM reply@)
**Answer**:
room: livingroom_0
entrance: door_2, doorway_3, doorway_9
stair: stairs_10
object: table_4, chair_5, sofa_6, window_13, tv_16, stool_17, couch_18, remote_19
\end{lstlisting}
\end{llmpromptbox}

\subsection{State Estimator prompts}

\begin{llmpromptbox}[specstyle, label=listing:place_name_struct]{Templated prompt for Place Label Similarity}
\begin{lstlisting}[style=Prompt]
**Question**: In the scene, there are ["kitchen_1", "livingroom_1", "livingroom_2", "bedroom_1", "familyroom_2"]. Now, I need to find a "livingroom". Please give me all the |<PlaceClass>| with similar semantic meaning from the list. Please directly return the answer in a Python list. Follow the format: Answer: <your answer>." 
\end{lstlisting}
\end{llmpromptbox}

\begin{llmpromptbox}[specstyle, label=listing:place_name_ex]{Example interaction with LLM for Place Label Similarity}
\begin{lstlisting}[style=Prompt]
@@Example query@@
**Question**: In the scene, there are ["kitchen_1", "livingroom_1", "livingroom_2", "bedroom_1", "familyroom_2"]. Now, I need to find a "livingroom". Please give me all the rooms with similar semantic meaning from the list. Please directly return the answer in a Python list. Follow the format: Answer: <your answer>." 

(@Example LLM reply@)
Answer: livingroom_1, livingroom_2, familyroom_2.
\end{lstlisting}
\end{llmpromptbox}

\begin{llmpromptbox}[specstyle, label=listing:place_recognition_struct]{Templated prompt for Pairwise Place Matching}
\begin{lstlisting}[style=Prompt]
@@Context@@
You are a robot exploring an environment for the first time. You will be given an object to look for and should provide guidance on where to explore based on a series of observations. Observations will be given as descriptions of objects seen from four cameras in four directions. Your job is to estimate the robot's state. You will be given two descriptions, and you need to decide whether these two descriptions describe the same |<PlaceClass>|. For example, if we have visited the |<PlaceClass>| before and got one description, when we visit a similar |<PlaceClass>| and get another description, it is your job to determine whether the two descriptions represent the same |<PlaceClass>|. You should understand that descriptions may contain errors and noise due to sensor noise and partial observability. Always provide reasoning along with a deterministic answer. If there are no suitable answers, leave the space after 'Answer: None.' Always include: Reasoning: <your reasoning> Answer: <your answer>.

@@Few-shot prompts@@
(*@\textbf{<Few-shot prompts follow format of query shown below. Examples of few-shot prompts given in \autoref{listing:place_recognition_ex}>}@*)

@@Query@@
**Description1**: I can see a brown wood headboard, ...
**Description2**: I can see a white wooden door, ...
**Question**: These are depictions of what I observe from two different vantage points. Please assess the shared objects and spatial relationship in the descriptions to determine whether these two positions are indeed in the same place. Provide a response of True or False, along with supporting reasons. In each direction, focus on only two to three large objects for reasoning.
**Answer**: _
**Reasoning**: _
\end{lstlisting}
\end{llmpromptbox}

\begin{llmpromptbox}[specstyle, label=listing:place_recognition_ex]{Example interaction with LLM for Pairwise Place Matching}
\begin{lstlisting}[style=Prompt]
@@Context@@
You are a robot exploring an environment for the first time...

@@Few-shot prompts@@
**Description1**: On the left, I can see a brown wood headboard, white paper pillow. On the right, I can see a black metal television, gray plastic laundry basket, white wood closet dresser, brown wood drawer. In fron of me, I can see a white wood bed, white wood window, brown metal lamp, brown wood dresser, brown wood dresser nightstand, black silk curtain, white plastic curtain, white metal wall lamp, brown wood drawer. Behind me, I can see a brown wood cabinet. 
**Description2**: On the left, I can see a white wood door. On the right, I can see a white wood bed, white glass lamp, white glass window, white plastic curtain, brown wood dresser nightstand, white glass window, white wood nightstand, blue fabric curtain, white cotton pillow, white metal ceiling fan, silver metal wall lamp. In front of me, I can see a white glass lamp, brown wood headboard, white cotton pillow, brown wood dresser, white wood bed, white cotton pillow, red metal wall lamp door, brown wood drawer. Behind me, I can see a brown wood bureau, black glass television, brown wood stool, brown wood drawer, brown wood drawer, brown wood drawer.
**Question**: These are depictions of what I observe from two different vantage points. Please assess the shared objects and spatial relationship in the descriptions to determine whether these two positions are indeed in the same place. Provide a response of True or False, along with supporting reasons. In each direction, focus on only two to three large objects for reasoning.
**Answer**: True
**Reasoning**: To simplify the description, given the abundance of objects, we initially focus on common perceptions of object sizes. Our attention is directed toward larger objects, as these are less prone to detection errors.
Description 1: On the left, there is brown wood headboard, white paper pillow. On the right, black metal television, brown wood dresser, and white wood closet dresser are of relatively large size. In front of me, there is a white wood bed and a brown wood dresser. Behind me, I can see a brown wood cabinet and black silk curtain.
Description 2: On the left, there is white wood door. On the right, the white wood bed, brown wood dresser, nightstand, and ceiling fan are of relatively large size. In front of me, there is a white wood bed, black glass television a brown wood dresser. Behind me, I can see a brown wood cabinet.
Shared Large Objects: the two descriptions exhibit significant commonalities, prominently featuring large and easily observable items such as a brown wood dresser, a brown wood drawer, a white wood bed, a metal wall lamp, a television, and a brown wood headboard. Spatial Relationship: the spatial relationships within both descriptions remain consistent, with the dresser and wall lamp positioned near the bed in each scenario. Despite minor variations in the color or material of smaller objects like stools or curtains, these discrepancies appear more likely to stem from observational nuances rather than indicating distinct rooms. 

...

@@Example query@@
**Description1**: On the left, there is a silver metal faucet, white glass mirror, black metal wall lamp, white stainless steel sink, white formica countertop, blue plastic accessory, white wood white door, silver metal faucet, white metal wall lamp, brown wood cabinet, black wood white door doorway, brown wood cabinet, white wood mirror wall. On the right, there is a brown tile floor, white drywall ceiling. In front of me, there is a silver metal faucet, silver glass mirror, white soap soap, white porcelain sink, brown wood cabinet, white porcelain sink, silver metal wall lamp, silver metal wall lamp. At the rear, there is an orange glass lamp, brown wood floor, white porcelain tub, white cotton bed.
**Description2**:  On the left, there is a white glass mirror, silver metal faucet, white porcelain sink, white wood bathroom sink, white wood bed, brown metal wall lamp. On the right, there are no specified items. In front of me, there is a silver metal faucet, white porcelain sink, white plastic toiletry, silver glass mirror, blue plastic soap toiletry, white white bathroom sink countertop, blue metal wall lamp, black wood bathroom cabinet, black metal wall lamp. Behind me, there is a white porcelain tub, black cloth curtain, white cotton bed, brown wood bed, black cloth curtain, white porcelain bath.
**Question**: These are depictions of what I observe from two different vantage points. Please assess the shared objects and spatial relationship in the descriptions to determine whether these two positions are indeed in the same place. Provide a response of True or False, along with supporting reasons. In each direction, focus on only two to three large objects for reasoning.

(@Example LLM reply@)
**Answer**: True
**Reasoning**: To simplify the description, given the abundance of objects, we focus initially on common perceptions of larger objects, as they are less prone to detection errors.
Description 1: On the left, there is brown wood cabinet, silver metal faucet, white wood bathroom sink, and white wood bed. On the right, there is nothing. In front of me, there is a silver metal faucet, silver glass mirror,white porcelain sink, brown wood cabinet, black mental wall lamp.  Behind me, there is an orange glass lamp, white porcelain tub, white cotton bed. 
Description 2: On the left, there is white porcelain sink and white wood bed. On the right, there are nothing. In front of me, there is a silver mental faucet and white plastic toiletry. Behind me, I can see white porcelain tub, black cloth curtain, white cotton bed.
Shared Large Objects: Both descriptions mention silver metal faucets, white beds, and sink, contributing to a consistent thematic presence. Spatial Relationship: The placement of key items like silver metal faucets and brown wood cabinets described in similar configurations create a sense of coherence. Potential for Small Errors: Description 2 lacks specified items, such as accessory, door and countertop. Besides, there are small errors in the material of bed. Errors in material or color, as well as the potential omission of small-size objects, can indeed be attributed to variations in observation. Despite this, the two descriptions are depicting the same room, with the potential for small errors accounting for minor variations.
\end{lstlisting}
\end{llmpromptbox}

\subsection{\graphshort{} Updater prompts}

\begin{llmpromptbox}[specstyle, label=listing:data_assoc_ex]{Example interaction with LLM for Object Data Association}
\begin{lstlisting}[style=Prompt]
@@Few-shot prompts@@
**Object to be matched**: We want to find a doorframe that is near a tv, a chair and a stool.
**List of observed objects**: Now we have seen the following object: doorframe_2 that is near chair and sofa. doorframe_3 that is near a tv and a chair. wooden door_2 that is near table, sink and lamp.
**Question**: Please select one object that is most likely to be the object I want to find. Please only select one object in the list and use this element name in answer. Use the exact name in the given sentences. Always follow the format: Answer: <your answer>."
**Answer**: doorframe_3
**Reasoning**: Among the given objects, "doorframe_3" is mentioned to be near a TV and a chair, most likely meeting the specified criteria of being near a TV, chair, and stool.

...

@@Example query@@
**Object to be matched**: We want to find a door that is near a dining table, window glass and a table cloth.
**List of observed objects**: Now we have seen the following object: door_2 that is near table and window glass. doorframe_4 that is near a table cloth and a glass. door_4 that is near photo, sofa and windows.
**Question**: Please select one object that is most likely to be the object I want to find. Please only select one object in the list and use this element name in answer. Use the exact name in the given sentences. Always follow the format: Answer: <your answer>."

(@Example LLM reply@)
**Answer**: door_2
**Reasoning**: Among the given objects, 'door_2' is mentioned to be near a table and window glass. Although 'doorframe_4' is also near a tablecloth and a glass, it seems to meet the criteria. However, since the target goal is a door and not a doorframe, 'door_2' is more suitable.
\end{lstlisting}
\end{llmpromptbox}

\begin{llmpromptbox}[specstyle, label=listing:region_abs_struct]{Templated prompt for Inferring Region Abstractions}
\begin{lstlisting}[style=Prompt]
Previously, we were in |<AbstractionClass>| containing the |<PlaceClass>|, and we move towards |<Subgoal>|. Now we arrive in a |<PlaceClass>|. Do you think the current state |<PlaceClass>| belongs to [list of existing |<AbstractionClass>|s]?

If it belongs to any other existing region abstraction, return the region abstraction name; otherwise propose the name for this new |<AbstractionClass>| formatted as "<name> |<AbstractionClass>| (New)". Ensure that your response follows the format:
Reasoning: <your reasoning>. 
Answer: <your answer>
\end{lstlisting}
\end{llmpromptbox}

\begin{llmpromptbox}[specstyle, label=listing:region_abs_ex]{Example interaction with LLM for Inferring Region Abstractions}
\begin{lstlisting}[style=Prompt]
@@Example query@@
Previously, we were in the Cooked Food Section containing the Deli Aisle, and we move towards Freezer. Now we arrive in a Dairy Aisle. Do you think the current state Dairy Aisle belongs to any of the existing region abstractions: ["Cooked Food Section", "Clothing Section", "Dry Goods Section"]? 

If it belongs to any other existing region abstraction, return the region abstraction name; otherwise propose the name for this new section formatted as "<name> Section (New)". Ensure that your response follows the format: 
Reasoning: <your reasoning>. 
Answer: <your answer>

(@Example LLM reply@)
Reasoning: The Dairy Aisle typically contains products such as milk, cheese, yogurt, and butter, which are perishable food items. Given its nature as a section primarily housing food products, it doesn't fit within the existing region abstractions provided ("Cooked Food Section", "Clothing Section", "Dry Goods Section").

Answer: Perishable Food Section (New)
\end{lstlisting}
\end{llmpromptbox}

\section{Reasoner details}
\label{app:reasoner_details}

\subsection{Region Proposer prompts}

\begin{llmpromptbox}[specstyle, label=listing:region_prompt_struct]{Templated prompt for Region Proposal}
\begin{lstlisting}[style=Prompt]
@@Few-shot prompts@@
(*@\textbf{<Few-shot prompts follow format of query shown below. Examples of few-shot prompts given in \autoref{listing:region_prompt_ex}>}@*)

@@Query@@
(*@\textbf{\graphlong{}}@*): You see the partial layout of the environment: {"room": {"livingroom_1": ...
**Goal**: Your goal is to find a |<Goal>| object.
**Question**: If any of the |<LayerType>| in the layout are likely to contain the target object, specify the most probable |<LayerType>| name. If all the |<LayerType>| are not likely to contain the target object, provide the |<ConnectorClass>| (that could connect <LayerType>) you would select for exploring a new |<LayerType>| where the target object might be found.
**Answer**: _
**Reasoning**: _
\end{lstlisting}
\end{llmpromptbox}

\begin{llmpromptbox}[specstyle, label=listing:region_prompt_ex]{Example interaction with LLM for Region Proposal}
\begin{lstlisting}[style=Prompt]
@@Few-shot prompts@@
(*@\textbf{\graphlong{}}@*): 
You see the partial layout of the apartment: {"room": {"kitchen_1": {"connects to": ["stair_1"]}, "livingroom": {"connects to": ["stair_1"]}, "entrance": {"stair_1": {"is near": []}}}}
**Goal**: Your goal is to find a sink.
**Question**: If any of the rooms in the layout are likely to contain the target object, specify the most probable room name. If all the room are not likely contain the target object, provide the door you would select for exploring a new room where the target object might be found.
**Answer**: kitchen_1
**Reasoning**: There are kitchen and livingroom in the layout. Among all the rooms, kitchen is usually likely to contain sink. Since we haven't explored the kitchen yet, it is possible that the sink is in the kitchen. Therefore, I will explore kitchen.

(*@\textbf{\graphlong{}}@*): 
You see the partial layout of the apartment: {"room": {"livingroom_1", "connects to": ["door_1", "door_2"]}, "diningroom_1": {,"connects to": ["door_1"]}}, "entrance": {"door_1": {"is near": ["towel_1"], "connects to": ["livingroom_1", "diningroom_1"]}, "door_2": {"is near": [], "connects to": ["livingroom_1"]}}}
**Goal**: Your goal is to find a sink.
**Question**: If any of the rooms in the layout are likely to contain the target object, specify the most probable room name. If all the room are not likely contain the target object, provide the door you would select for exploring a new room where the target object might be found.
**Answer**: door_1
**Reasoning**: There is only livingroom in the layout. livingroom is not likely to contain sink, so I will not explore the current room. Among all the doors, door1 is near to towel. A towel is usually more likely to near the bathroom or kitchen, so it is likely that if you explore door1 you will find a bathroom or kitchen and thus find a sink.

...

@@Example query@@
(*@\textbf{\graphlong{}}@*): 
You see the partial layout of the environment: {"room": {"livingroom_1": {"connects to": ["doorway_1", "door_2"]}, "entrance": {"doorway_1": {"is near": ["table_1"]}, "door_2": {"is near": ["clock_1"], "connects to": ["livingroom_1" ]}}} 
**Goal**: Your goal is to find a oven.
**Question**: If any of the rooms in the layout are likely to contain the target object, specify the most probable room name. If all the room are not likely contain the target object, provide the door you would select for exploring a new room where the target object might be found.

(@Example LLM reply@)
**Answer**: doorway_1
**Reasoning**: There are only livingroom in the layout. Among all the rooms, livingroom is usually unlikely to contain an oven, making it less likely for me to find an oven in the current room. Instead, I plan to explore other rooms connected to the current living room via entrances. Evaluating the entrances, doorway1 stands out as it is close to a table. Tables are commonly found in kitchens, which often contain ovens. Therefore, I have decided to explore through doorway_1.
\end{lstlisting}
\end{llmpromptbox}

\subsection{Object Proposer prompts}

\begin{llmpromptbox}[specstyle, label=listing:object_prompt_struct]{Templated prompt for Object Proposal}
\begin{lstlisting}[style=Prompt]
@@Few-shot prompts@@
(*@\textbf{<Few-shot prompts follow format of query shown below. Examples of few-shot prompts given in \autoref{listing:object_prompt_ex}>}@*)

@@Query@@
**Objects in current Place node**: There is a list: ["mirror_2", "lamp_1", ...]. 
**Question**: Please select one object that is most likely located near a |<Goal>|. Always follow the format: Answer: <your answer>.
**Answer**: _
**Reasoning**: _
\end{lstlisting}
\end{llmpromptbox}

\begin{llmpromptbox}[specstyle, label=listing:object_prompt_ex]{Example interaction with LLM for Object Proposal}
\begin{lstlisting}[style=Prompt]
@@Few-shot prompts@@
**Objects in current Place node**:
There is a list: ["chair_4", "sofa_2", "bed_9", "dresser_1","ceiling_6","closet_5", "window_7", "wall_10"]. 
**Question**: Please select one object that is most likely located near a table. Always follow the format: Answer: <your answer>. Reasoning: <your Reasoning>.
**Answer**: chair_4
**Reasoning**: Among the given options, the object most likely located near a table is a "chair." Chairs are commonly placed around tables for seating during various activities such as dining, working, or socializing.
...

@@Query@@
**Objects in current Place node**:
There is a list: ["mirror_2", "lamp_1", "picture_7", "tool_6","toilet_8","sofa_11", "floor_12", "wall_13"].
**Question**: Please select one object that is most likely located near a sink. Always follow the format: Answer: <your answer>. Reasoning: <your Reasoning>.

(@Example LLM reply@)
**Answer**: mirror_2
**Reasoning**: Among the given options, the object most likely located near a sink is a "mirror." Mirrors are commonly found near sinks in bathrooms for personal grooming and hygiene activities.
\end{lstlisting}
\end{llmpromptbox}

\section{Automatic \graphshort{} generation details}
\subsection{Description generation prompts}
\begin{llmpromptbox}[specstyle, label=listing:auto_gen_description_struct]{Templated prompt for Environment Description Generation}
\begin{lstlisting}[style=Prompt]
@@Context@@
You are a helpful system. Given a environment type, you are required to describe the environment layout based on the meta structure.
In meta structure, we define 4 abstract types of classes to describe the environment. 
1. Objects. Static scene elements occupying spatially localised regions: e.g. tables, chairs etc. They provide semantic cues for reasoning, and act as navigation goals and localisation landmarks - specifically, we distinguish Objects/Places with object features, aggregated lists of nearby objects. 
2. Places. The smallest semantically meaningful spatial regions in a scene, e.g. rooms in a home or aisles in a supermarket, which typically cannot be subdivided in a consistent, meaningful way. 
3. Region Abstractions. Spatial abstractions encompassing multiple smaller spatial regions: e.g. a floor containing multiple rooms, or a supermarket section that is a cluster of aisles.
4. Connectors. These are specific types of Objects that are also structural scene elements connecting spatial regions: e.g. doors connecting rooms, steps connecting floors.

@@Query@@
We are using text to describe the environment layout. Please follow the meta-structure to describe the environment. 

Based on the example, please return the text to describe |<EnvironmentLabel>|'s layout. Object is always required. Follow the response format: Text: <your text>.
\end{lstlisting}
\end{llmpromptbox}

\begin{llmpromptbox}[specstyle, label=listing:auto_gen_description_ex]{Example interaction with LLM for Environment Description Generation}
\begin{lstlisting}[style=Prompt]
@@Context@@
You are a helpful system. Given a environment type...

@@Query@@
We are using text to describe the environment layout. Please follow the meta-structure to describe the environment. 

Example:

Environment Type: household

Text: In the household environments, there are often three layers including room, entrance and object. Home contains various rooms. The room contains spatial region ans is Places in meta structure. Rooms can be connected to rooms or entrance. Entrances are Connectors in meta structure. Rooms contains objects. Entrance and Objects is near objects.

Based on the example, please return the text to describe |<EnvironmentLabel>|'s layout. Object is always required. Follow the response format: Text: <your text>.

@@Feedback@@
The text you generate contains the error: |<MetastructureErrorMessage>|. Please fix this issue and re-generate the text to describe |<EnvironmentLabel>| layout. Object is always required. Follow the response format: Text: <your text>.
\end{lstlisting}
\end{llmpromptbox}

\subsection{Canonicalisation prompts}

\begin{llmpromptbox}[specstyle, label=listing:auto_gen_triplets_struct]{Templated prompt for Triplet Extraction}
\begin{lstlisting}[style=Prompt]
@@Few-shot prompts@@
(*@\textbf{<Few-shot prompts follow format of query shown below. Examples of few-shot prompts given in \autoref{listing:auto_gen_triplets_ex}>}@*)

@@Query@@
We use triplets to structure the text. Triplets follow the template: [Subject, Relation, Object].

Example:
Text: In the household environments, there are often three layers including room, entrance and object. Home contains various rooms. The room contains spatial region ans is Places in meta structure. Rooms can be connected to rooms or entrance. Entrances are Connectors in meta structure. Rooms contains objects. Entrance and Objects is near objects.
Triplets:
[room, contains, object]
[room, connects, entrance]
[room, connects, room]
[entrance, is near, object]

Based on the example, please return the triplets of |<EnvironmentLabel>| environment. Only allow |<MetastructureRelationTypes>| these relations. Follow the format: Triplets: <your triplets>. 
Text: <text_description>.
\end{lstlisting}
\end{llmpromptbox}

\begin{llmpromptbox}[specstyle, label=listing:auto_gen_triplets_ex]{Example interaction with LLM for Triplet Extraction}
\begin{lstlisting}[style=Prompt]
@@Few-shot prompts@@
**Question:** We use triplets to structure the text. Triplets follow the template: [Subject, Relation, Object].
Given the description of household environment: In the household environments, there are often three layers including room, entrance and object. Home contains various rooms. The room contains spatial region ans is Places in meta structure. Rooms can be connected to rooms or entrance. Entrances are Connectors in meta structure. Rooms contains objects. Entrance and Objects is near objects.
Please return the triplets of the household environment. Follow the format: Triplets: <your triplets>.

**Triplets:**
[room, contains, object]
[room, connects, entrance]
[room, connects, room]
[entrance, is near, object]

@@Query@@
We use triplets to structure the text. Triplets follow the template: [Subject, Relation, Object].

Example:
Text: In the household environments, there are often three layers including room, entrance and object. Home contains various rooms. The room contains spatial region ans is Places in meta structure. Rooms can be connected to rooms or entrance. Entrances are Connectors in meta structure. Rooms contains objects. Entrance and Objects is near objects.
Triplets:
[room, contains, object]
[room, connects, entrance]
[room, connects, room]
[entrance, is near, object]

Based on the example, please return the triplets of |<EnvironmentLabel>| environment. Only allow |<MetastructureRelationTypes>| these relations. Follow the format: Triplets: <your triplets>. 
\end{lstlisting}
\end{llmpromptbox}

\begin{llmpromptbox}[specstyle, label=listing:auto_gen_canonicalisation_struct]{Templated prompt for Triplet Canonicalisation}
\begin{lstlisting}[style=Prompt]
@@Query@@
Given a peice of text and a relational triplet extracted from it, and the definition of the relation in it, choose the most appropriate relation to replace it in this context if there is any and modify the triplet. If the entity is too specific, use the abstract word to replace it.
In the |<EnvironmentLabel>|, |<Abstraction1>| |<Relation>| |<Abstraction2>|. Triplet: (|<Abstraction1>|, |<Relation>|, |<Abstraction2>|)

We define 4 abstract types of classes to describe the environment. 
1. Objects. Static scene elements occupying spatially localised regions: e.g. tables, chairs etc. They provide semantic cues for reasoning, and act as navigation goals and localisation landmarks - specifically, we distinguish Objects/Places with object features, aggregated lists of nearby objects. 
2. Places. The smallest semantically meaningful spatial regions in a scene, e.g. rooms in a home or aisles in a supermarket, which typically cannot be subdivided in a consistent, meaningful way. 
3. Region Abstractions. Spatial abstractions encompassing multiple smaller spatial regions: e.g. a floor containing multiple rooms, or a supermarket section that is a cluster of aisles.
4. Connectors. These are specific types of Objects that are also structural scene elements connecting spatial regions: e.g. doors connecting rooms, steps connecting floors.

For subject entity or object entity, you should first decide the abstract class that the entity belongs to and make sure the entity name is at the right abstraction level for |<EnvironmentLabel>|. 
If the subject entity or object entity is too specific, use an abstract word to replace it. For example, if products are a specific kind of object and are not Connectors, we should replace them with "objects". If the object is a Connector that connects regions, such as entrances, we should not replace it with "objects". 
Do not use any class type name, including "Places", "Region Abstractions", and "Connectors", as the entity name in triplets except for "objects".

We define 4 relations as follows.
1. "contains": Specifies spatial containment relations among regions/objects. The subject entity should contain the object entity. The subject entity is required to be a region or a place. Example: [floor, contains, room], ['section', 'contains', 'aisle']
2. "connects to": Specifies reachability between the subject entity and object entity, and can connect Place, Region Abstraction, and Connectors. Object cannot "connects to" any other items. Example: [room, connects, entrance], [room, connects to, room]
3. "is near": It approximately capture closeness between the subject entity and object entity. The subject entity must be Connector and the object entity must be objects. Example: [entrance/door, is near, object]
4. "invald": If the relation is none of the above, it is invalid. For example, the object cannot be near a place or a region.

For example, in household environment, given [place, has, desk], we should modify it to [room, contains, object]. In household environment, given [connector, connects to, rooms], we should modify it to [entrance, connects to, room]. 

Please follow the format: Answer: <modified triplets>.
\end{lstlisting}
\end{llmpromptbox}

\section{System details}
\label{app:system_details}
\subsection{System hyperparameters}

In \autoref{tab:hyperparams}, we list the various modules in the system, describe the hyperparameters used in each and provide the values used in \navsys{} for our experiments. The \engineshort{}'s and Reasoner's hyperparameters are kept the same across both simulation and real-world tests, and across robot platforms in the real world. Some of the thresholds are sensor-dependent and can benefit from tuning for a specific sensor - \ie{} centroid nearness and minimum object size thresholds that are expressed in pixel-based units, and hence depend on image resolution or field-of-view. In practice, our centroid nearness threshold is defined to be large enough that sufficient object features can be accumulated when using both RGB sensors in simulation and on our robots in the real world.

\begin{table*}
    \caption{\textbf{Hyperparameters used in the \navsys{} system.}}
    \begin{center}
        \begin{tabular}{@{\extracolsep{4pt}}lll@{}}
        \toprule
        \textbf{Subsystem} & \textbf{Hyperparameter} & \textbf{Value}  \\
        \midrule
        \multirow{4}{*}{\begin{tabular}{l}
            \engineshort{}
        \end{tabular}}& Centroid nearness \centroidthresh{} (object nearness heuristic) & 100 (pixels) \\
        & Bounding box overlap \bboxthresh{} (object nearness heuristic) & 0.1 (IoU)  \\
        & Min. object size threshold & 200 (pixels$^{2}$) \\
        & LLM temperature & 0.3 \\
        \midrule
        \multirow{1}{*}{\begin{tabular}{l}
            Reasoner
        \end{tabular}}& LLM temperature & 0.3 \\
        \midrule
        \multirow{5}{*}{\begin{tabular}{l}
            Controller\\
            (Simulation - FMM)
        \end{tabular}}& Occupancy map resolution & 0.05 (m) \\
        & Occupancy map size & 4.8 (m) \\
        & Min. obstacle height & 0.5 (m) \\
        & Turning action angle & 30 (deg) \\
        & Forward action distance & 0.25 (m) \\
        \midrule
        \multirow{3}{*}{\begin{tabular}{l}
             Controller\\
             (Real-world - GNM)
        \end{tabular}}& Stopping threshold & 2.25 \\
        & Max linear velocity & 0.5 (m/s) \\
        & Max angular velocity & 0.3 (rad/s) \\
        \bottomrule
        \end{tabular}
    \label{tab:hyperparams}
    \end{center}
\end{table*}

\section{Additional experimental results}

\subsection{Experimental results and details for \graphshort{} mapping}
\label{app:exp_mapper_rw}
\begin{specbox}[specstyle, label=listing:osg_spec_supermarket]{Supermarket}
\begin{lstlisting}[language=json]
`Aisle`:
    ``layer_type``: "Place"
    ``layer_id``: 2
    ``contains``: [Object]
    ``connects_to``: [Aisle]
`Object`:
    ``layer_id``: 1
\end{lstlisting}
\end{specbox}

\begin{figure*}
    \centering
    \includegraphics[width=\textwidth]{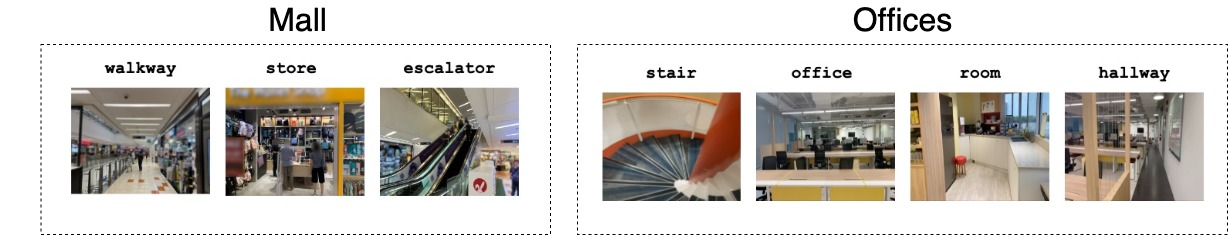}
    \caption{\textbf{Spatial concepts from Mall/Office \graphshort{} specs.} Full \graphshort{} specifications using these concepts are given in \autoref{listing:osg_spec_office} and \autoref{listing:osg_spec_mall}.}
    \label{fig_app:sg_rw_elements}
\end{figure*}

\begin{specbox}[specstyle, label=listing:osg_spec_office]{Office}
\begin{lstlisting}[language=json]
`Floor`:
    ``layer_type``: "Region"
    ``layer_id``: 3
    ``contains``: [Hallway, Office, Room]
    ``connects_to``: [Stairs]
`Hallway`:
    ``layer_type``: "Place"
    ``layer_id``: 2
    ``contains``: [Object]
    ``connects_to``: [Hallway, Office, Room, Entrance]
`Office`:
    ``layer_type``: "Place"
    ``layer_id``: 2
    ``is_near``: [Object]
    ``connects_to``: [Hallway, Office, Room, Entrance]
`Room`:
    ``layer_type``: "Place"
    ``layer_id``: 2
    ``is_near``: [Object]
    ``connects_to``: [Hallway, Office, Room, Entrance]
`Stair`:
    ``layer_type``: "Connector"
    ``layer_id``: 2
    ``is_near``: [Object]
    ``connects_to``: [Floor]
`Entrance`:
    ``layer_type``: "Connector"
    ``layer_id``: 2
    ``is_near``: [Object]
    ``connects_to``: [Hallway, Office, Room]
`Object`:
    ``layer_id``: 1
\end{lstlisting}
\end{specbox}

\begin{specbox}[specstyle, label=listing:osg_spec_mall]{Mall}
\begin{lstlisting}[language=json]
`Floor`:
    ``layer_type``: "Region"
    ``layer_id``: 3
    ``contains``: [Walkway, Store]
    ``connects_to``: [Escalator]
`Walkway`:
    ``layer_type``: "Place"
    ``layer_id``: 2
    ``contains``: [Object]
    ``connects_to``: [Walkway, Store, Escalator]
`Store`:
    ``layer_type``: "Place"
    ``layer_id``: 2
    ``is_near``: [Object]
    ``connects_to``: [Store, Walkway]
`Escalator`:
    ``layer_type``: "Connector"
    ``layer_id``: 2
    ``is_near``: [Object]
    ``connects_to``: [Floor, Walkway]
`Object`:
    ``layer_id``: 1
\end{lstlisting}
\end{specbox}

\begin{figure*}
    \centering
    \includegraphics[width=\textwidth]{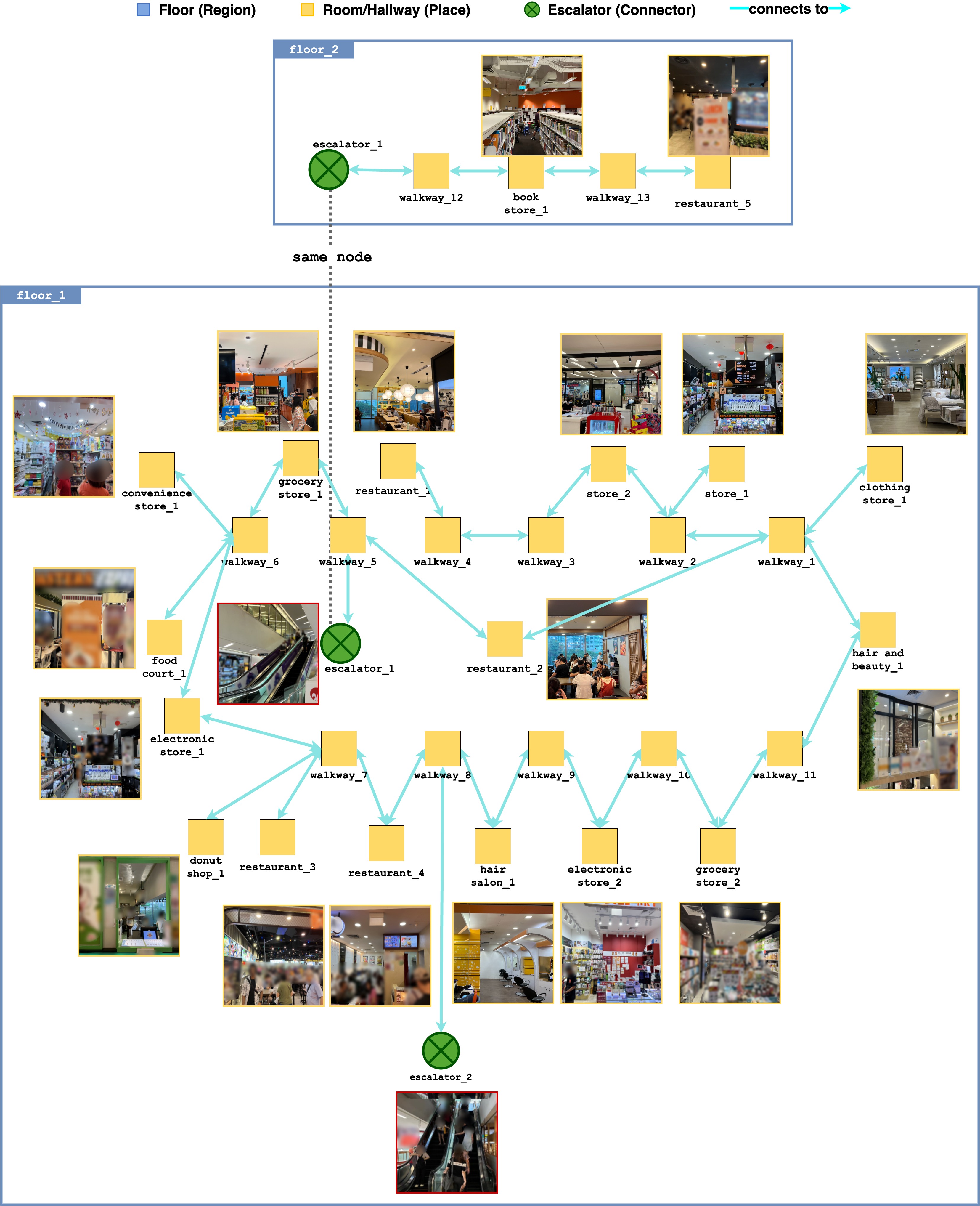}
    \caption{\textbf{\graphshort{} of real-world Mall environment.} The spatial concepts used to describe this \textit{walkways} and \textit{stores} for Places, and \textit{escalators} for Connectors. We provide the full \graphshort{} specification in \autoref{listing:osg_spec_mall}.}
    \label{fig_app:sg_rw_mall}
\end{figure*}

\end{document}